\documentclass{article}

\usepackage[preprint]{neurips_2026}

\usepackage[utf8]{inputenc} 
\usepackage[T1]{fontenc}    
\usepackage{hyperref}       
\usepackage{url}            
\usepackage{booktabs}       
\usepackage{amsfonts}       
\usepackage{nicefrac}       
\usepackage{microtype}      
\usepackage{xcolor}         

\usepackage{amsmath}
\usepackage{comment}
\usepackage{graphicx}
\usepackage{natbib}

\usepackage{booktabs}
\usepackage{multirow}

\usepackage[most]{tcolorbox}
\usepackage{listings}
\usepackage{enumitem}

\usepackage{xcolor}
\usepackage{enumitem}
\usepackage{titlesec}
\usepackage{hyperref}
\usepackage{orcidlink}
\usepackage{makecell}

\setcitestyle{numbers,square}

\title{Argument Graph Uncertainty: Quantifying Uncertainty from the Logical Structure of Reasoning Chains}

\author{%
    Simon Baur\orcidlink{0009-0009-4307-3078} \\
    Fraunhofer Heinrich-Hertz-Institut\\
    10587 Berlin, Germany \\
    \texttt{simon.baur@hhi.fraunhofer.de} \\
    \And
    Sumire Honda \orcidlink{0009-0008-5166-5885}\\
    Fraunhofer Heinrich-Hertz-Institut\\
    10587 Berlin, Germany \\
    \texttt{sumire.honda@hhi.fraunhofer.de} \\
    \And
    Wojciech Samek \orcidlink{0000-0002-6283-3265}\\
    Fraunhofer Heinrich-Hertz-Institut\\
    10587 Berlin, Germany \\
    \texttt{wojciech.samek@hhi.fraunhofer.de} \\
    \And
    Jackie Ma \orcidlink{0000-0002-2268-1690} \\
    Fraunhofer Heinrich-Hertz-Institut\\
    10587 Berlin, Germany \\
    \texttt{jackie.ma@hhi.fraunhofer.de} \\
}

\begin{document}
\maketitle

\begin{abstract}
 Reliable uncertainty estimation is essential for deploying large language models in high-stakes reasoning tasks, yet most existing approaches rely on expensive sampling strategies or shallow token-level signals that ignore the internal logical structure of model outputs. We introduce Argument Graph Uncertainty, a post-hoc framework for multiple choice question tasks, that estimates model confidence directly from the logical structure of a single chain-of-thought reasoning trace. Our method uses either a lightweight instruction-tuned model or a simple deterministic rule to segment the reasoning chain into propositional units, then applies a NLI model to infer relational edges and construct two complementary argument graphs over these segments. The resulting edge scores are aggregated into a Dirichlet posterior over answer options, whose concentration yields calibrated uncertainty estimates reflecting both local step-to-step coherence and global logical consistency across the reasoning trace. Evaluated on GPQA, MMLU-Pro and TruthfulQA, our method consistently surpasses token-based and consistency-based uncertainty baselines across multiple reasoning models. We further demonstrate through extensive combination analyses that argument graph uncertainty captures an orthogonal signal to existing approaches, making it a complementary component in uncertainty ensembles. Crucially, our framework requires no additional calls to the evaluated model and relies entirely on small, efficient auxiliary models, producing fully interpretable, per-node uncertainty attributions at low computational cost.
\end{abstract}

\section{Introduction}
\label{sec:intro}

\begin{figure}
  \centering
  \includegraphics[width=\linewidth]{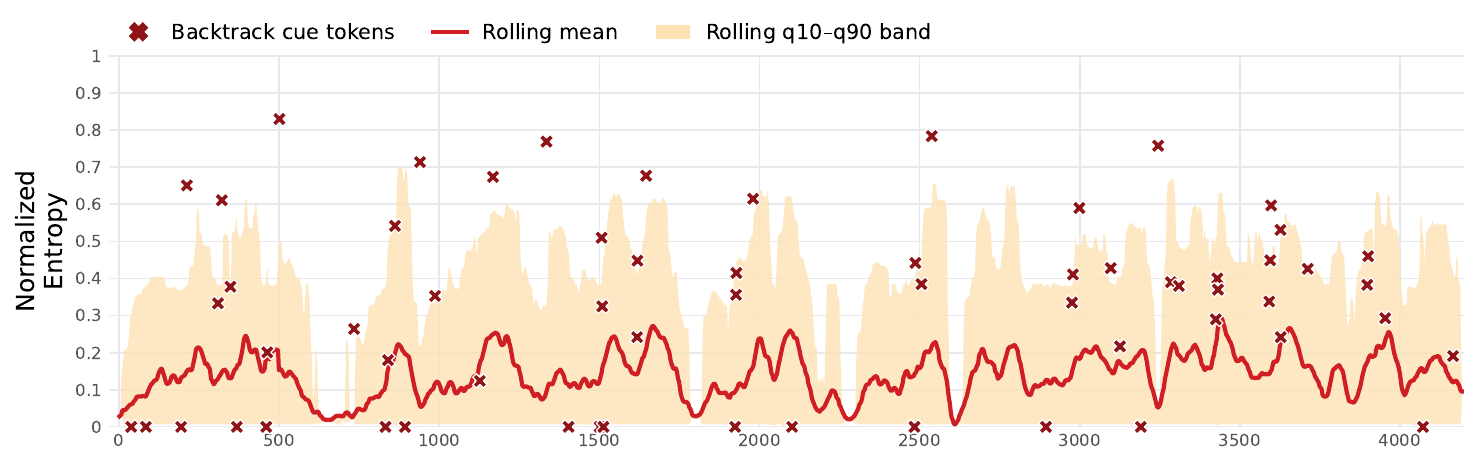}
  \caption{\textbf{Token-level entropy across reasoning trace of a single GPQA example.} Normalized per-token entropy (rolling mean, q10--q90 band) and backtracking cue tokens ({\color{red}$\times$}) across a single reasoning trace. The vast majority of tokens are generated with high confidence, with intermittent spikes, causing the rolling mean to settle roughly between ${\approx}0.1 - {\approx}0.2$. Backtracking cues exhibit strongly varying entropy, and do not align with spikes or plateaus. Reasoning traces of correct and incorrect predictions yield similar token entropy trajectories, making it unclear which tokens to select or how to aggregate them into a meaningful uncertainty signal.}
  \label{fig:token_entropy}
\end{figure}

The landscape of artificial intelligence has been fundamentally reshaped by the emergence of reasoning models \citep{jaech2024openai, guo2025deepseek}. Unlike traditional LLMs that rely on immediate token prediction, these models engage in deliberate, multi-step computation before committing to an answer. As these systems move from scientific performance benchmarks to high-stakes deployments and autonomous agentic loops, the need for additional metrics, in particular, robust uncertainty quantification (UQ) becomes critical. A safe deployment in such domains (e.g. medical diagnosis or legal decisions) hinges not only on a model's accuracy, but on its ability to signal when its own internal reasoning process has failed to converge on a stable or logically consistent conclusion~\citep{abdar2021review,xia2025survey}. Existing UQ methods for language models fall into two camps, and both are ill-suited to reasoning models. \textbf{Consistency-based methods} sample multiple outputs and measure their agreement~\citep{lin2023generating,kuhn2023semantic}. For reasoning models generating thousands of tokens per trace, this multiplies inference cost by an order of magnitude, making the approach impractical in many deployment settings. Moreover, by looking only at the final-answer agreement, these methods discard all information in the reasoning trace itself. \textbf{Token-level entropy methods} use token log-probabilities as a confidence proxy~\citep{fadeeva2024fact,gupta2024language}. These are computationally less expensive, but the signal degrades precisely where it is needed most~\citep{entropytrajectory2025}, as the deeper context mechanically inflates token confidence regardless of the reasoning quality. Simpler heuristics like counting trace length \cite{devic2025trace}, backtracking phrases \cite{kumar2024training}, or peak-entropy tokens \cite{xiong2023can} seemed to offer an easy and inexpensive way to catch uncertainty signals from reasoning traces, however recent work uncovered major shortcomings of these approaches: Trace length is confounded by problem difficulty, RL-induced length bias~\citep{yuan2025efficient,nohara2026optimal}, overthinking~\citep{hassid2025don}, and the apparent signal in overthinking tokens has been traced to training artifacts rather than genuine uncertainty~\citep{notthinkingstraight2025}. Token-entropy also cannot separate semantic uncertainty (genuine ambiguity about the answer) from irrelevant lexical variation. Figure \ref{fig:token_entropy} gives an intuition of noisy and incoherent behavior token uncertainty can exhibit when confronted with complex tasks.
\begin{figure}
  \centering
  \includegraphics[width=\linewidth]{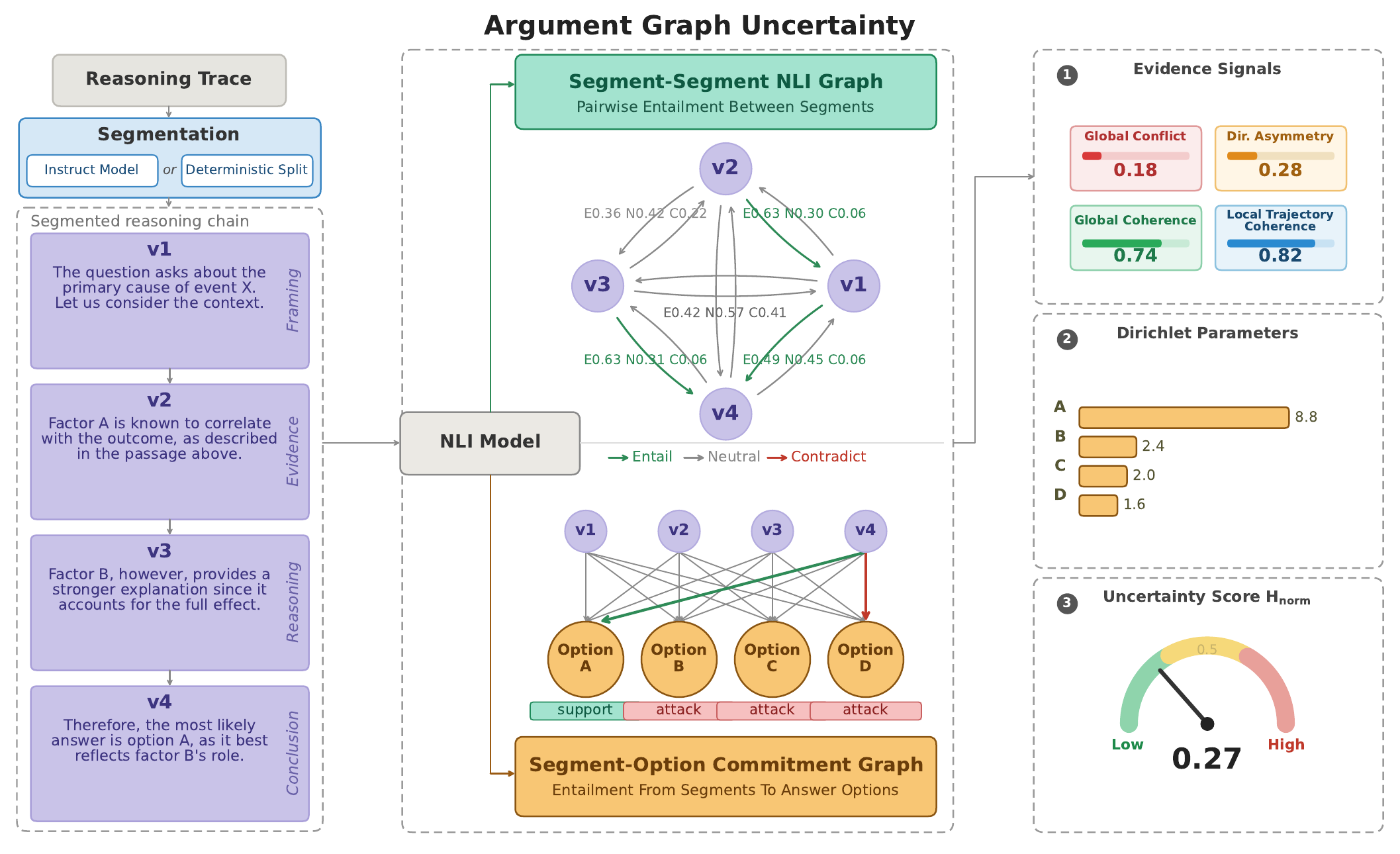}
  \caption{\textbf{Argument Graph Uncertainty.} A reasoning trace is segmented into propositional nodes $\{v_1, \dots, v_4\}$ and passed through an NLI model to construct two complementary graphs: a \emph{Segment--Segment Graph} capturing pairwise entailment, conflict, and asymmetry across all nodes, and a \emph{Segment--Option Commitment Graph} mapping each node to answer options. Graph-derived evidence signals are aggregated into Dirichlet concentration parameters $\{\alpha_o\}$, from which a normalized predictive entropy $H_{\mathrm{norm}}$ serves as the final uncertainty score.}
  \label{fig:graph_uncertainty}
\end{figure}
Finally, token- and consistency-based approaches are further compounded by entropy collapse ~\citep{cui2025entropymechanismreinforcementlearning,liu2025uncertainty, jin2025revisiting}: RL training sharpens output distributions as models exploit high-reward trajectories, making token-level signals increasingly unstable and suppressing the diversity that consistency methods depend on.
The core issue is that all of these approaches ignore what actually determines whether a reasoning chain should be trusted: its \emph{logical structure}. What premises are introduced? Do intermediate steps support or contradict one another? Does the conclusion follow from well-grounded inferential chains? None of this is captured by surface statistics or consistency aggregations. Current research in UQ lacks a framework capable of assessing these internal dynamics. To fill this gap, we propose \textit{Argument Graph Uncertainty} (AGU), a post-hoc UQ framework that estimates model confidence from the logical structure of a single chain-of-thought trace for Multiple-Choice-Question-Answering (MCQA). AGU segments the trace into propositional units via a small instruction-tuned model. It then employs a lightweight natural language inference (NLI) model to infer the support and attack relations between these units, constructing two complementary argument graphs that capture global consistency, local trajectory coherence, and node-to-option evidence. By mapping these structural signals into a Dirichlet posterior, AGU provides a calibrated uncertainty estimate that remains robust even when traditional probabilistic signals fail. Figure \ref{fig:graph_uncertainty} provides a graphical overview of individual steps of our proposed method. We show that AGU consistently outperforms token-based and consistency-based baselines across multiple reasoning models for the GPQA, MMLU-PRO and TruthfulQA benchmarks~\citep{rein2023gpqa, wang2024mmlu,lin2022truthfulqa}. We further empirically show that AGU captures an orthogonal signal, making it a natural complement to existing methods. Furthermore, it adds only small computational overhead compared to consistency based methods, since it relies entirely on small auxiliary models. Our contributions can be summarized as follows:

\begin{itemize}
    \item \textbf{A Novel Reasoning Trace-Based Framework for UQ:} We propose a new method for uncertainty quantification specifically designed for reasoning models. By exploiting the logical trajectories of reasoning traces, our approach captures structural uncertainty signals that are absent in established methods.
    \item \textbf{Outperforming established Methods:} We demonstrate that our method surpasses both token- and consistency-based uncertainty quantification techniques.
    \item \textbf{Orthogonality of Graph-based UQ signals :} Through detailed empirical combination analysis we show that graph-derived uncertainty serves as an orthogonal signal to consistency based approaches. This allows our method to be integrated with already existing UQ methods, and provide a stronger measure of uncertainty than either single method would on its own.
\end{itemize}

\section{Related Work}
\label{sec:related_work}

\paragraph{Uncertainty Quantification for LLMs}
\citet{xia-etal-2025-survey} categorized existing UQ approaches on LLMs into four: \textit{verbalizing methods}, \textit{latent information
methods}, \textit{consistency-based methods}, and \textit{semantic clustering methods}.
Verbalizing methods prompt a model to output uncertainty in natural language with the prediction \cite{lin2022teaching}. 
Although the approach is straightforward and intuitive, instruction models' poor performance in expressing their uncertainty has been observed \citep{yona-etal-2024-large}. 
Latent information methods use probability distributions on each generated token, which we refer to as \textit{token-level entropy}. 
They typically estimate uncertainty using specific measures, such as averaged or maximum of negative log-likelihood \cite{manakul-etal-2023-selfcheckgpt} or entropy of the response distribution in single or multiple responses \cite{portillo-wightman-etal-2023-strength}.
Consistency-based methods measure consistency or variance over multiple responses based on similarity \citep{cole-etal-2023-selectively}. 
Several metrics are proposed to estimate similarity, from lexical-based metrics such as BLEU \citep{papineni-etal-2002-bleu, Huang_2025} to more sophisticated approaches capturing the semantic level, such as SentenceBERT or NLI-based methods \citep{gao-etal-2024-spuq, wang2024subjectiveuncertaintyquantificationcalibration}.
Semantic clustering methods also consider multiple responses, however instead of measuring similarity scores, they cluster the response by their semantic equivalence and calculate the probability over the clusters to estimate uncertainty \citep{kuhn2023semantic}.
They employ NLI classifier or general LLMs to determine the entailment relationships  \citep{kuhn2023semantic, farquhar2024detecting}.
Our approach relates to semantic clustering methods, however it differs in applying clustering methods in reasoning traces in terms of the logical relevance to the categorical answer options using argument graphs.

\paragraph{Uncertainty Quantification for Reasoning Models}
\citet{pmlr-v238-harsha-tanneru24a} proposed \textit{Chain of Thought Uncertainty}, which measures the agreement between two Chain-of-Thought (CoT) explanations by employing an NLI classifier \citep{he2021deberta}.
They found a strong correlation between confidence in CoT explanations and the correctness of the final responses, highlighting the importance of incorporating CoT into UQ.
On the other hand, \citet{agarwal2026madelivingbenchmarkmultilabel} observed surprisingly weak UQ performance especially on large reasoning models, leaving a critical gap in reliable uncertainty estimation for reasoning-intensive tasks.
\citet{zhang2025cotuqimprovingresponsewiseuncertainty} distilled critical information by extracting keywords and assessed their degree of importance to the final answer.
\citet{ballon2026probingtrajectoriesreasoningtraces} investigated how accuracy evolves over the reasoning trajectory by partitioning CoT traces into deciles and truncating them at each cutoff point.
On GPQA Diamond \citep{rein2023gpqa} and MMLU-PRO \citep{wang2024mmluprorobustchallengingmultitask}, which require multi-step reasoning, they observed a progressive score increase across successive deciles that aligned semantically with reasoning progress, rather than merely reflecting context length or structural form.
However, for long-horizon reasoning chains, the issue of inflated uncertainty scores must be addressed \citep{korbak2025chainthoughtmonitorabilitynew}; to this end, \citet{mao2026recurrentconfidencechaintemporalaware} proposed a recurrent propagation mechanism that yields more accurate uncertainty estimates by integrating historical confidence information across steps.
\citet{da2025understandinguncertaintyllmexplanations} introduced a method for incorporating structural information from reasoning chains by applying a graph edit distance-based measure.

\paragraph{Abstract Argumentation Frameworks.} Our approach is conceptually grounded in the work of \citet{dung1995acceptability}, which introduced the \emph{Abstract Argumentation Framework}. Dung's framework treats arguments as atomic nodes in a directed graph where edges represent "attack" relations. The "acceptability" of a set of arguments is then determined by graph-theoretic properties such as conflict-freedom and admissibility. While Dung's work focused on formal logic programming and $n$-person games, it provides the mathematical foundation for identifying consistent subsets of reasoning steps. We adapt this intuition to reasoning chains of LLMs, moving from binary attack relations to probabilistic NLI-based scores that quantify the degree of conflict and support within a generated trace.

\paragraph{NLI Modelling}
Natural Language Inference is the task of determining whether a hypothesis sentence is entailed, contradicted, or neutral, given a premise \citep{farquhar_detecting_2024, bowman-etal-2015-large}. 
Given the relevance to the logical support and oppositions between propositions, NLI therefore provides a mechanism to infer relational edges, which we employ in our approach. 
We use DeBERTa \citep{he2021debertadecodingenhancedbertdisentangled} as pretrained backbone to infer entailment, neutrality or contradictions that serve as edges in the argument graphs we construct.  
DeBERTA further enables computationally efficient modeling of dependency between premise-hypothesis pairs. 
DeBERTa model remains the strongest encoder-based model on NLI benchmarks despite the relatively low computational cost \citep{laurer2024buildingefficientuniversalclassifiers}. 

\paragraph{Evidential Deep Learning and Dirichlet Posteriors.} 
To map our graph-based signals onto a formal uncertainty signal, we adopt a Evidential Deep Learning (EDL) framework. Following \citet{sensoy2018evidential}, we treat the parameters of a Dirichlet distribution as evidence for the predicted class probabilities. While EDL has been extensively utilized to quantify epistemic uncertainty and detect out-of-distribution (OOD) samples in computer vision \citep{malinin2018predictive} and subjective logic for NLP classification \citep{josang2016subjective}, its application to reasoning structures remains unexplored. AGU therefore represents a departure from classical EDL: rather than deriving evidence from raw output logits, we compute it from the structural integrity and logical coherence of the model's argumentation graph.

\paragraph{Multiple-Choice Question Answering }
Multiple-choice question answering (MCQA) is a natural evaluation paradigm for calibrated uncertainty, as discrete selection over a fixed candidate set mirrors real-world decision problems including agentic tool selection~\citep{patil2024gorilla}, retrieval-augmented re-ranking, and structured clinical triage~\citep{kanjee2023accuracy}. The closed-form answer space affords unambiguous accuracy measurement, providing a clean signal for evaluating calibration quality~\citep{ohuncertainty, duan2025uprop, zhou2026orch}. We adopt GPQA Diamond~\citep{rein2023gpqa} as our general benchmark. Its difficulty ensures that memorization alone cannot explain performance, as evidenced by the limited accuracy achieved by state-of-the-art systems ~\citep{rein2023gpqa}. We further include MMLU-PRO and TruthfulQA to asses generalizability across broad domains ~\citep{wang2024mmlu,lin2022truthfulqa}.

\section{Method}
\label{sec:method}

AGU is a post-hoc uncertainty estimation framework that treats a model's chain-of-thought as an \emph{argument graph}. By doing so, it exploits the internal logical structure of a single generated reasoning trace. It decomposes the reasoning chain into discrete propositional segments, constructs two complementary relational graphs over those segments using a NLI scorer, and aggregates the resulting signals into a Dirichlet posterior over answer options whose concentration is then used to infer entropy that serves as uncertainty estimate.

\subsection{Reasoning Segmentation}
\label{sec:segmentation}
 
Given a model-generated chain-of-thought $R$ for a multiple-choice question with $K$ answer options $\mathcal{A} = \{a_1, \dots, a_K\}$, our goal is to arrive at an uncertainty score $U_{\mathrm{AGU}}(R)$. To this end, we partition $R$ into $n$ textual segments $\mathcal{V} = \{v_1, \dots, v_n\}$ using a light-weight instruct tuned LLM, where each $v_i$ corresponds to a semantically coherent reasoning step. Throughout, $i, j \in \{1, \dots, n\}$ index segments and $k \in \{1, \dots, K\}$ indexes answer options.

\subsection{NLI Edge Scoring}
\label{sec:nli}

All relational edges are then scored by a pretrained NLI model $f_{\mathrm{NLI}}$. For a premise $p$ and a hypothesis $h$, the model returns a probability simplex over three classes:
\begin{equation}
    f_{\mathrm{NLI}}(p, h) = \bigl(p_E(p, h),\; p_N(p, h),\; p_C(p, h)\bigr),
    \qquad p_E + p_N + p_C = 1,
\end{equation}
where $E$, $N$, $C$ denote entailment, neutral, and contradiction, respectively. The first argument is always the premise and the second the hypothesis. The NLI model is applied to two types of input pairs: \emph{segment--segment} pairs $(v_i, v_j)$, with segment $v_i$ as premise and segment $v_j$ as hypothesis (Section~\ref{sec:all_to_all}), and \emph{segment--option} pairs $(v_i, h_k)$, with segment $v_i$ as premise and the hypothesis $h_k$ = \emph{``The answer is Option X: [option text]''} constructed from answer option $a_k$ as hypothesis (Section~\ref{sec:node_option}), for which we write $p_E(v_i, h_k)$, $p_N(v_i, h_k)$, $p_C(v_i, h_k)$ analogously.

\subsection{Node--Option Graph}
\label{sec:node_option}

For each segment $v_i \in \mathcal{V}$ and each answer option $a_k \in \mathcal{A}$, we define the \textbf{contradiction-adjusted support} of node $v_i$ for option $a_k$ as:
\begin{equation}
    e_{i,k} = \max\!\bigl(0,\; p_E(v_i, h_k) - p_C(v_i, h_k)\bigr).
    \label{eq:node_evidence}
\end{equation}
This quantity is strictly positive only when a node's entailment of an option outweighs its contradiction of that option, i.e.\ it counts as evidence for an option only if it supports that option more than it contradicts it. Rectifying at zero means a step arguing against an option contributes nothing, rather than cancelling evidence contributed by other steps.

\subsection{All-to-All Segment Graph}
\label{sec:all_to_all}

For every ordered pair $(v_i, v_j)$ with $i \neq j$, we score $v_i$ as the premise and $v_j$ as the hypothesis, yielding $p_E(i,j)$, $p_N(i,j)$, $p_C(i,j)$. This \emph{segment-to-segment} graph captures global logical consistency across the full reasoning chain. We extract two scalar properties from the segment-to-segment graph. We define \textbf{global conflict} as the mean pairwise contradiction across all directed edges:
\begin{equation}
    \mathrm{Conflict} = \frac{1}{n(n-1)} \sum_{i \neq j} p_C(v_i, v_j).
    \label{eq:conflict}
\end{equation}
Global conflict measures how much the trace argues against itself, averaged over all scored segment pairs. High conflict indicates that reasoning steps are mutually inconsistent. We define \textbf{directional asymmetry} as the mean absolute difference in entailment scores between opposing directions:
\begin{equation}
    \mathrm{Asymmetry} = \frac{2}{n(n-1)} \sum_{i < j} \bigl|p_E(v_i,v_j) - p_E(v_j,v_i)\bigr|.
    \label{eq:asym}
\end{equation}

Directional asymmetry measures whether the entailment between two steps survives swapping premise and hypothesis. High asymmetry indicates that the support between nodes depends on their order, i.e.\ that a step follows from another without the reverse holding. Scoring all $n(n-1)$ ordered pairs is quadratic in the number of segments. Alternatively, we can restrict the segment-to-segment graph to pairs within a window of width $w$ along the reasoning chain, $|i - j| \le w/2$, which reduces the number of scored pairs to ${\approx}\,n\,w$. Each segment is thus compared only with its $w/2$ preceding and $w/2$ following segments, in both directions,
so the graph captures local rather than global consistency. Conflict, asymmetry and the global coherence (see Section~\ref{sec:node_weights}) are then averaged over the scored pairs instead of all pairs; the remaining steps are unchanged.

\subsection{Node Reliability Weights}
\label{sec:node_weights}
Each node $v_i$ receives a reliability weight combining both global and local graph signals. The raw coherence (\textbf{global coherence}) of node $v_i$ over the all-to-all graph is its mean net entailment across all other nodes:

\begin{equation}
    \mathrm{coh}^{\mathrm{raw}}_i = \frac{1}{n-1} \sum_{j \neq i} \bigl(p_E(v_i,v_j) - p_C(v_i,v_j)\bigr) \in [-1, 1], \quad \text{normalized as} \quad \mathrm{coh}_i = \tfrac{1}{2}(\mathrm{coh}^{\mathrm{raw}}_i + 1).
    \label{eq:coh}
\end{equation}

We define \textbf{local trajectory coherence} as follows: let $\mathcal{T}(i)$ denote the set of adjacent edges incident to $v_i$, i.e.\ $(v_{i-1} \to v_i)$ and $(v_i \to v_{i+1})$, each scored with the earlier segment as premise and the later one as hypothesis; the first and last segment have only one such edge. The raw local coherence is:
\begin{equation}
    \mathrm{traj}^{\mathrm{raw}}_i = \frac{1}{|\mathcal{T}(i)|} \sum_{j \in \{i-1,\, i\}} \bigl(p_E(v_j, v_{j+1}) - p_C(v_j, v_{j+1})\bigr) \in [-1, 1],
    \label{eq:traj}
\end{equation}
normalized as $\mathrm{traj}_i = \tfrac{1}{2}(\mathrm{traj}^{\mathrm{raw}}_i + 1)$.
We then compute a \textbf{final node weight} as a weighted sum of global and local coherence:
\begin{equation}
    r_i = \lambda \,\mathrm{coh}_i + (1-\lambda)\,\mathrm{traj}_i,
    \label{eq:node_weight}
\end{equation}
where $\lambda \in [0, 1]$ is a hyperparameter. Equations~\eqref{eq:coh}--\eqref{eq:node_weight} weight each step by how well it fits the trace as a whole
(global coherence, Eq.~\eqref{eq:coh}) and its immediate neighbours (local coherence,
Eq.~\eqref{eq:traj}): a step is trusted to the extent that it is consistent with both, with
$\lambda$ setting the balance (Eq.~\eqref{eq:node_weight}).

\subsection{Dirichlet Evidence Aggregation}
\label{sec:dirichlet}

We then aggregate node-level evidence into a per-option signal:
\begin{equation}
    E_k = \sum_{i=1}^{n} r_i \, e_{i,k}, \qquad k = 1, \dots, K.
    \label{eq:weighted_evidence}
\end{equation}
To translate this evidence into a distribution over options, we adopt a Dirichlet model~\citep{sensoy2018evidential}. The effective evidence scale is penalized by the global graph irregularities identified above:
\begin{equation}
    \kappa_{\mathrm{eff}} = \kappa \exp\!\left(-\gamma_c\,\mathrm{Conflict} - \gamma_a\,\mathrm{Asymmetry}\right),
    \label{eq:kappa_eff}
\end{equation}
where $\kappa$, $\gamma_c$, $\gamma_a > 0$ are hyperparameters. Chains with high internal conflict or directional instability thus receive a globally attenuated evidence scale, reflecting reduced trustworthiness of the reasoning trace as a whole. We can then go on to compute Dirichlet parameters as $\alpha_k = \alpha_0 + \kappa_{\mathrm{eff}}\, E_k, \; k = 1, \dots, K$, where $\alpha_0 > 0$ is a symmetric prior. The predictive mean over options can be derived as:
\begin{equation}
    \hat{p}_k = \frac{\alpha_k}{S}, \qquad S = \sum_{k=1}^{K} \alpha_k.
    \label{eq:pred_mean}
\end{equation}

\subsection{Uncertainty Quantification}
\label{sec:uncertainty}

We compute uncertainty estimates from the Dirichlet posterior as Shannon entropy over the predictive mean, normalized to $[0, 1]$:

\begin{equation}
    U_{\mathrm{AGU}}(R)=H_{\mathrm{norm}} = \frac{-\sum_{k=1}^{K} \hat{p}_k \log \hat{p}_k}{\log K}.
    \label{eq:entropy}
\end{equation}
High entropy indicates that probability mass is spread broadly across options, while low entropy reflects a concentrated, confident prediction.

\section{Experimental Setting and Results}
\label{sec:results}
\paragraph{Experimental Setting}
We evaluate our framework on the aforementioned benchmarks, using nine publicly available reasoning models: DeepSeek-R1-Llama-70B, DeepSeek-R1-Qwen-14B, DeepSeek-R1-Llama-8B, GPT-OSS-120B, GPT-OSS-20B, Nemotron-Nano-30B, Qwen3-30B-Thinking, Qwen3-4B-Thinking, and Phi-4-Mini-Reasoning. All models are run in FP16 precision with sampling temperature $0.7$ and 
top-$p$ $0.9$. We report average metrics over $n{=}5$ sampled answers and reasoning traces per question. We use Qwen-8B Instruct for deriving segments $\mathcal{V}$ from reasoning traces if not stated otherwise. We attach the prompt used for segmentation in Appendix \ref{appendix_c_prompt}. For detailed description of how scores and metrics are defined see Appendix \ref{appendix:a_experimental_settings}.  All experiments reported in Table \ref{tab:main_table} are conducted with hyperparameters $\alpha_0 = 1.0$, $\kappa = 12.0$, $\gamma_{\mathrm{conflict}} = 1.2$, $\gamma_{\mathrm{asymm}} = 1.0$ and $\lambda =0.5$. Appendix \ref{appendix_hyper_params} offers a thorough analysis on hyperparameter influences. While our chosen default set of parameters is sufficient to achieve good performance, some models could benefit from hyperparameter tuning, as shown in moderate increases of PRR and $\rho$.

\paragraph{Uncertainty Calibration}

Table~\ref{tab:main_table} reports uncertainty calibration across all nine reasoning models for GPQA diamond. 
$U_{\mathrm{AGU}}$ achieves the highest PRR in eight out of nine models, demonstrating 
consistent gains over both $U_{\mathrm{Cons}}$ and $U_{\mathrm{Token}}$ across model families 
and scales. The improvements are particularly pronounced for smaller models such as 
Phi-4 Mini Reasoning and Qwen3 4B Thinking, where token-level uncertainty proves 
unreliable and consistency-based methods underperform, suggesting that argument-graph 
structure captures uncertainty signals that are otherwise lost. The single exception is 
DeepSeek-R1-D.-Llama-70B, where $U_{\mathrm{Cons}}$ attains the highest PRR 
($0.46 \pm 0.09$). Token uncertainty $U_{\mathrm{Token}}$ 
performs the least reliably overall, with negative PRR values for GPT-OSS 120B and 
DeepSeek-R1-D.-Llama-70B indicating that token probabilities are miscalibrated to the 
point of inverting the rejection ordering, in line with the critical stance we took in Section \ref{sec:intro}. Appendix \ref{appendix:additional_benchmarks} provides calibration results for MMLU-Pro and TruthfulQA. 

\begin{figure}[!h]
  \centering
  
  \includegraphics[width=\linewidth]{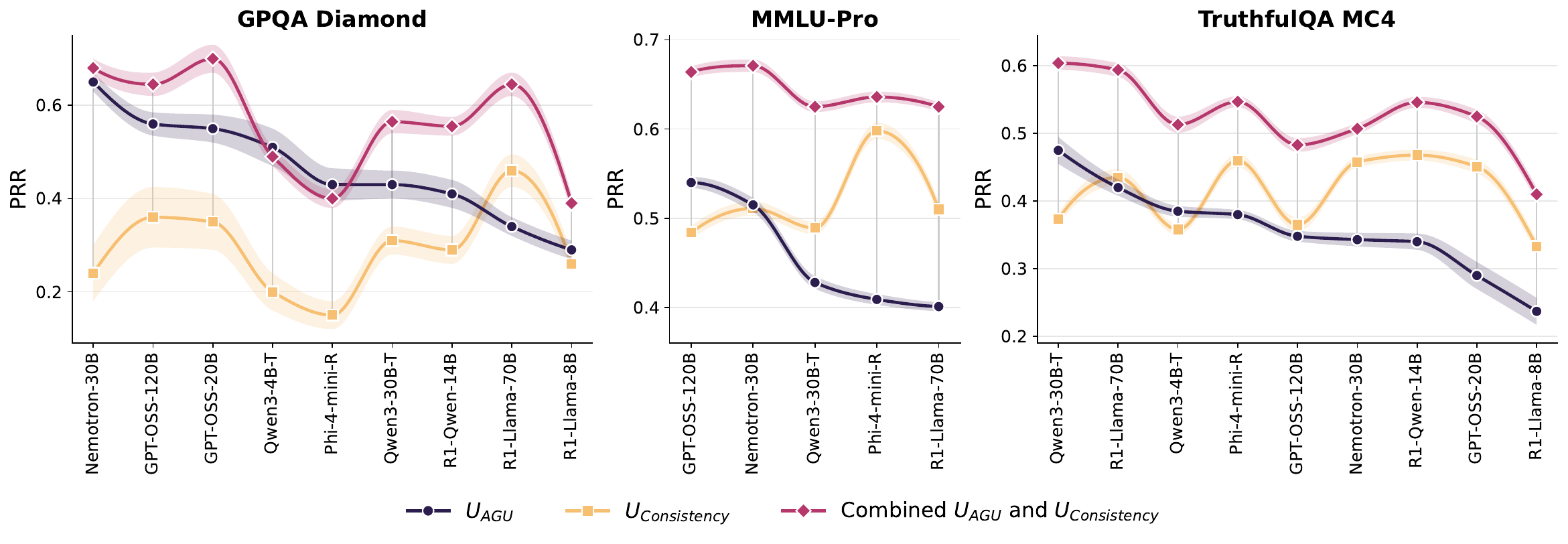}
  \caption{\textbf{PRR across all models for $U_{\mathrm{AGU}}$, $U_{\mathrm{Cons}}$, 
    and their equal-weight combination.} The combined score consistently dominates both 
    individual methods.}
    \label{fig:combined}
\end{figure}
\begin{table*}[!h]
\centering
\small
\setlength{\tabcolsep}{4pt}
\caption{\textbf{Calibration performance across nine reasoning models on GPQA Diamond.} We compare Argument Graph Uncertainty $U_{\mathrm{AGU}}(R)$ against consistency-based uncertainty $U_{\mathrm{Cons}}$ and token probability uncertainty $U_{\mathrm{Token}}$ using Prediction Rejection Rate (PRR~$\uparrow$) and Spearman correlation ($\rho$~$\downarrow$). Bold indicates the best method per model. $U_{\mathrm{AGU}}(R)$ achieves the highest PRR in eight out of nine models.}
\label{tab:main_table}
\resizebox{\textwidth}{!}{%
\begin{tabular}{lcccccc}
\toprule
& \multicolumn{2}{c}{\textbf{Argument Graph Uncertainty $\boldsymbol{U_{\mathrm{AGU}}(R)}$}} & \multicolumn{2}{c}{$\boldsymbol{U_{\mathrm{Cons}}}$} & \multicolumn{2}{c}{$\boldsymbol{U_{\mathrm{Token}}}$} \\
\cmidrule(lr){2-3}\cmidrule(lr){4-5}\cmidrule(lr){6-7}
\textbf{Model} & \textbf{PRR $\uparrow$} & \textbf{$\rho$ $\downarrow$} & \textbf{PRR $\uparrow$} & \textbf{$\rho$ $\downarrow$} & \textbf{PRR $\uparrow$} & \textbf{$\rho$ $\downarrow$} \\
\midrule
Nemotron-3 Nano 30B & \textbf{0.65 $\pm$ 0.07} & \textbf{-0.37 $\pm$ 0.06} & 0.24 $\pm$ 0.14 & -0.24 $\pm$ 0.08 & 0.41 $\pm$ 0.15 & -0.32 $\pm$ 0.08 \\
GPT-OSS 120B & \textbf{0.56 $\pm$ 0.09} & -0.28 $\pm$ 0.06 & 0.36 $\pm$ 0.15 & \textbf{-0.37 $\pm$ 0.08} & -0.27 $\pm$ 0.19 & 0.02 $\pm$ 0.08 \\
GPT-OSS 20B & \textbf{0.55 $\pm$ 0.10} & \textbf{-0.37 $\pm$ 0.07} & 0.35 $\pm$ 0.14 & -0.36 $\pm$ 0.09 & 0.14 $\pm$ 0.18 & -0.20 $\pm$ 0.09 \\
Qwen3 4B Thinking & \textbf{0.51 $\pm$ 0.07} & \textbf{-0.39 $\pm$ 0.06} & 0.20 $\pm$ 0.09 & -0.21 $\pm$ 0.07 & 0.27 $\pm$ 0.08 & -0.21 $\pm$ 0.07 \\
Qwen3 30B Thinking & \textbf{0.43 $\pm$ 0.09} & -0.27 $\pm$ 0.06 & 0.31 $\pm$ 0.11 & \textbf{-0.32 $\pm$ 0.07} & 0.10 $\pm$ 0.02 & -0.10 $\pm$ 0.02 \\
Phi-4 Mini Reasoning & \textbf{0.43 $\pm$ 0.07} & \textbf{-0.35 $\pm$ 0.06} & 0.15 $\pm$ 0.08 & -0.13 $\pm$ 0.07 & 0.06 $\pm$ 0.08 & -0.05 $\pm$ 0.07 \\
DeepSeek-R1-D.-Qwen-14B & \textbf{0.41 $\pm$ 0.08} & \textbf{-0.32 $\pm$ 0.06} & 0.29 $\pm$ 0.09 & -0.28 $\pm$ 0.07 & 0.10 $\pm$ 0.10 & -0.14 $\pm$ 0.07 \\
DeepSeek-R1-D.-Llama-70B & 0.34 $\pm$ 0.04 & -0.24 $\pm$ 0.03 & \textbf{0.46 $\pm$ 0.09} & \textbf{-0.45 $\pm$ 0.06} & -0.39 $\pm$ 0.07 & 0.28 $\pm$ 0.06 \\
DeepSeek-R1-D.-Llama-8B & \textbf{0.29 $\pm$ 0.08} & \textbf{-0.27 $\pm$ 0.06} & 0.26 $\pm$ 0.08 & -0.22 $\pm$ 0.07 & -0.13 $\pm$ 0.08 & 0.07 $\pm$ 0.07 \\
\bottomrule
\end{tabular}%
}
\end{table*}

\begin{figure}[!h]
  \centering
  \includegraphics[width=\linewidth]{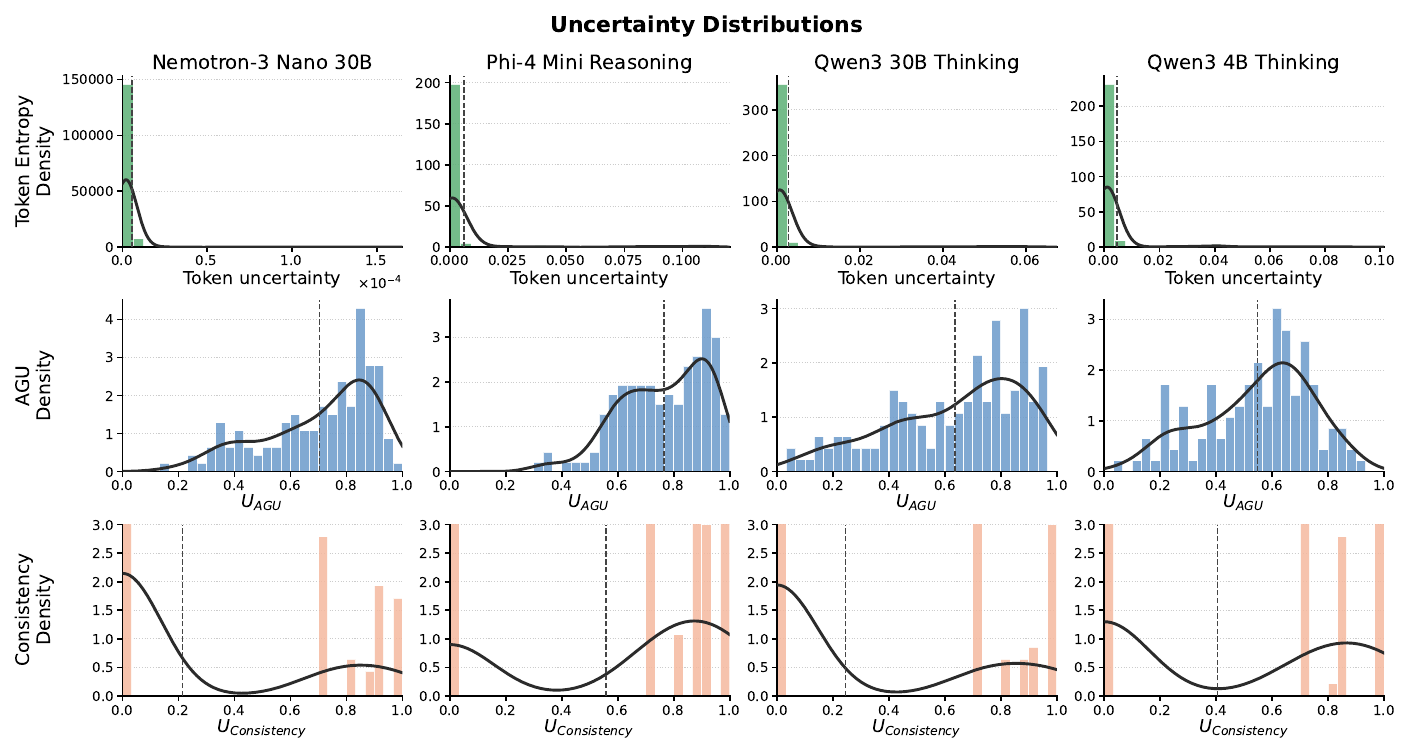}
     \caption{\textbf{Uncertainty score distribution examples.} Token entropy 
    (\textbf{top}) collapses to near-zero across all models, rendering it uninformative. 
    $U_{\mathrm{Cons}}$ (\textbf{bottom}) is strongly bimodal (models either agree 
    unanimously or disagree maximally) with little resolution in between. 
    $U_{\mathrm{AGU}}$ (\textbf{middle}) spreads broadly across $[0,1]$, capturing diverse 
    uncertainty signals and filling precisely the range that consistency cannot resolve.}
    \label{fig:histograms}
\end{figure}

Figure \ref{fig:combined} shows the effect of combining $U_{\mathrm{AGU}}$ and 
$U_{\mathrm{Cons}}$ via equal-weight blending. The combined score achieves the highest 
PRR in every model, and the gains are substantial: for Nemotron-3 Nano 30B the combined 
score reaches $0.68$, compared to $0.65$ for $U_{\mathrm{AGU}}$ alone and $0.24$ for 
$U_{\mathrm{Cons}}$ alone, and for DeepSeek-R1-D.-Llama-70B, the one model where 
$U_{\mathrm{Cons}}$ outperforms $U_{\mathrm{AGU}}$ individually, the combination 
still improves over both. This consistent pattern strongly suggests that the two 
signals are largely orthogonal: argument-graph structure captures uncertainty arising 
from internal reasoning conflicts, while consistency-based methods reflect variability 
across independent samples, and these failure modes do not overlap. Appendix \ref{combination_ablation} provides an extensive analysis of alternative 
combination strategies, offering the insight that basic equal blending is enough to utilize the potential of combining the two uncertainty signals. Furthermore, through visual inspection of distributions of uncertainty score we can get a sense of how well and why they might behave well or ill-calibrated: figure \ref{fig:histograms} provides an overview over token distributions of different methods for different models. Token uncertainty collapses to near-zero across all models, rendering it uninformative. 
Consistency scores are strongly bimodal: predictions are either unanimous or maximally 
uncertain, thereby offering little resolution in between. Argument graph uncertainty, by 
contrast, spreads broadly across $[0,1]$, providing a diverse and well-calibrated 
uncertainty signal.

\paragraph{Segmentation ablation.}
We ablate the Qwen3-8B segmenter used in the main experiments against two
alternatives: a deterministic split of the trace at paragraph boundaries, which
requires no model call, and a larger segmenter, GPT-OSS-120B
(Table~\ref{tab:segmentation_ablation}). The deterministic split loses no
calibration on average and gains it for several models, with mean PRR 0.47
against 0.46, and improves significantly on GPT-OSS 20B. The larger segmenter
yields the highest mean PRR at 0.50 and, notably, cuts the
standard deviation from 0.08 to 0.03 on average, so the segmentation stage is
the main source of run-to-run variance in $U_{\mathrm{AGU}}$. Since the paragraph split removes the segmentation cost
entirely (Eq.~\ref{eq:cost_agu}) at no loss in mean PRR, we consider it the
default choice when compute is a concern.

\begin{table*}[!h]
\centering
\small
\setlength{\tabcolsep}{3pt}
\caption{\textbf{Segmentation ablation on GPQA Diamond.} PRR of $U_{\mathrm{AGU}}$
across different segmenters, mean $\pm$ std over seeds. $^{\boldsymbol{*}}$ marks cells that
differ significantly from the Qwen3-8B row (Welch's $t$-test, $p<0.05$).}
\label{tab:segmentation_ablation}
\resizebox{\textwidth}{!}{%
\begin{tabular}{lccccccccc|c}
\toprule
\textbf{Segmenter} & \makecell{\textbf{Nemotron}\\\textbf{30B}} & \makecell{\textbf{GPT-OSS}\\\textbf{120B}} & \makecell{\textbf{GPT-OSS}\\\textbf{20B}} & \makecell{\textbf{Qwen3}\\\textbf{4B-T}} & \makecell{\textbf{Qwen3}\\\textbf{30B-T}} & \makecell{\textbf{Phi-4}\\\textbf{mini-R}} & \makecell{\textbf{R1-Qwen}\\\textbf{14B}} & \makecell{\textbf{R1-Llama}\\\textbf{70B}} & \makecell{\textbf{R1-Llama}\\\textbf{8B}} & \textbf{Mean} \\
\midrule
Qwen3-8B        & 0.65 $\pm$ .07 & 0.56 $\pm$ .09 & 0.55 $\pm$ .10 & 0.51 $\pm$ .07 & 0.43 $\pm$ .09 & 0.43 $\pm$ .07 & 0.41 $\pm$ .08 & 0.34 $\pm$ .04 & 0.29 $\pm$ .08 & 0.46 \\
GPT-OSS-120B    & 0.67 $\pm$ .07 & 0.68 $\pm$ .03$^{\boldsymbol{*}}$ & 0.76 $\pm$ .03$^{\boldsymbol{*}}$ & 0.39 $\pm$ .06$^{\boldsymbol{*}}$ & 0.47 $\pm$ .02 & 0.32 $\pm$ .03$^{\boldsymbol{*}}$ & 0.42 $\pm$ .01 & 0.46 $\pm$ .02$^{\boldsymbol{*}}$ & 0.31 $\pm$ .02 & 0.50 \\
Paragraph split & 0.71 $\pm$ .05 & 0.62 $\pm$ .03 & 0.76 $\pm$ .01$^{\boldsymbol{*}}$ & 0.35 $\pm$ .01$^{\boldsymbol{*}}$ & 0.46 $\pm$ .01 & 0.29 $\pm$ .03$^{\boldsymbol{*}}$ & 0.37 $\pm$ .04 & 0.36 $\pm$ .04 & 0.29 $\pm$ .03 & 0.47 \\
\bottomrule
\end{tabular}%
}
\end{table*}

\paragraph{Robustness to NLI errors.}
\begin{figure*}[!t]
  \centering
  \includegraphics[width=\textwidth]{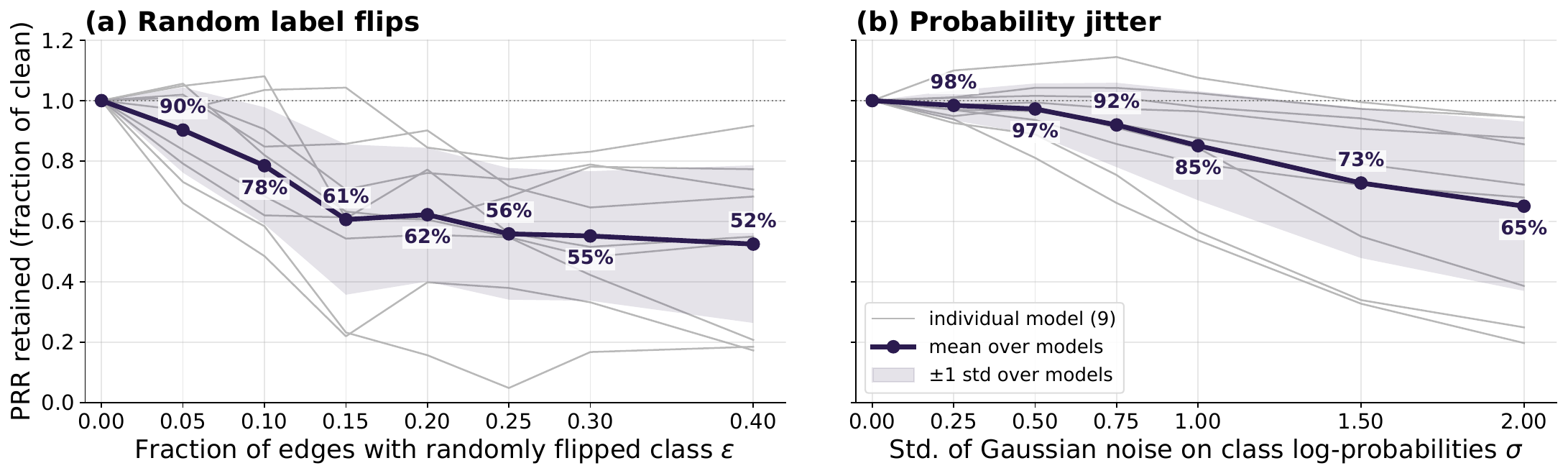}
  \caption{\textbf{Robustness of $U_{\mathrm{AGU}}$ to NLI errors.} PRR retained
  on GPQA Diamond when (a)~a fraction $\varepsilon$ of edge classes is flipped
  at random and (b)~Gaussian noise $\sigma$ is added to the class
  log-probabilities. Grey: individual models; bold: mean; band: $\pm$1 std.}
  \label{fig:nli_robustness}
\end{figure*}
Since $U_{\mathrm{AGU}}$ is computed from the judgments of a small NLI scorer,
we test how errors at the edge level propagate to the uncertainty estimate. We
perturb the scorer's outputs on all nine GPQA Diamond models and recompute
$U_{\mathrm{AGU}}$, either flipping the predicted class of a random fraction
$\varepsilon$ of edges or adding Gaussian noise of standard deviation $\sigma$
to the class log-probabilities (Fig.~\ref{fig:nli_robustness}). Degradation is
graceful in both cases. With 10\% of edges mislabelled, 78\% of the clean PRR
is retained, and half remains when 40\% of all edges carry a wrong class.
Noise on the probabilities costs almost nothing until it exceeds the typical
margin between classes, with 97\% retained at $\sigma{=}0.5$ and 85\% at
$\sigma{=}1$. Because the Dirichlet evidence aggregates over many edges, no
single misclassification moves the per-option estimate much, and
$U_{\mathrm{AGU}}$ depends more on the scorer's ordinal judgment rather than on
per-edge accuracy or calibration. Appendix~\ref{appendix:nli_robustness} gives
the full details. Appendix~\ref{appendix:edge_stats} provides additional analysis on distribution of edge classification by the NLI-scorer. 

\paragraph{All-to-all vs. windowed segment-to-segment graph.} As mentioned in \ref{sec:all_to_all} we can either use the full segment-to-segment (all-to-all) graph, or a windowed subset. We
provide an ablation of the all-to-all vs. the windowed graph in Appendix~\ref{appendix:window_ablation} showing that PRR
matches to within $0.01$ for a window of $w=10$ while the cost of the NLI stage reduces from quadratic
to linear in the number of segments. Therefore we suggest using the windowed approached as default.

\paragraph{AGU vs. trace length.}
Since longer traces contain more segments, we verify that $U_{\mathrm{AGU}}$
is not simply a proxy for trace length. On GPQA Diamond, the segment count used
directly as an uncertainty score attains a mean PRR of $0.33$, compared to
$0.45$ for $U_{\mathrm{AGU}}$, which is the stronger signal on seven of nine
models. The two are moderately correlated (mean Spearman $0.43$); after
rank-regressing the segment count out of $U_{\mathrm{AGU}}$, a PRR of $0.33$
remains, indicating that the majority of the graph signal is independent of
trace length. See Appendix\ref{appendix:length} for details.

\paragraph{Computational Cost}\label{sec:cost}
AGU adds no calls to the evaluated model; its overhead runs on an optional
segmenter with $P_{\mathrm{seg}}$ parameters and an NLI encoder with
$P_{\mathrm{NLI}}$ parameters. For a trace of $L$ tokens split into segments of
average length $\ell$, with $K$ answer options and window $w$,

\begin{equation}
  C_{\mathrm{AGU}} \;=\;
  \underbrace{%
  \begin{cases}
    P_{\mathrm{seg}}\,L \in \mathcal{O}(P_{\mathrm{seg}}\,L) & \text{\scriptsize LLM segmenter}\\[1pt]
    0 & \text{\scriptsize paragraph split}
  \end{cases}}_{\text{segmentation}}
  +
  \underbrace{%
  \begin{cases}
    2(w{+}K)\,P_{\mathrm{NLI}}\,L \in \mathcal{O}(P_{\mathrm{NLI}}\,L) & \text{\scriptsize window } w\\[1pt]
    2\,P_{\mathrm{NLI}}\,L^{2}/\ell \in \mathcal{O}(P_{\mathrm{NLI}}\,L^{2}/\ell) & \text{\scriptsize all-to-all}
  \end{cases}}_{\text{NLI scoring}}
  \label{eq:cost_agu}
\end{equation} 
A more detailed explanation and derivation are given in
Appendix~\ref{appedix:computational_costs}. Fig.~\ref{fig:computational_costs}
depicts these costs graphically alongside the consistency baseline and reports
measured wall-clock times on GPQA Diamond.

\begin{figure}[!h]
  \centering
  \includegraphics[width=\linewidth]{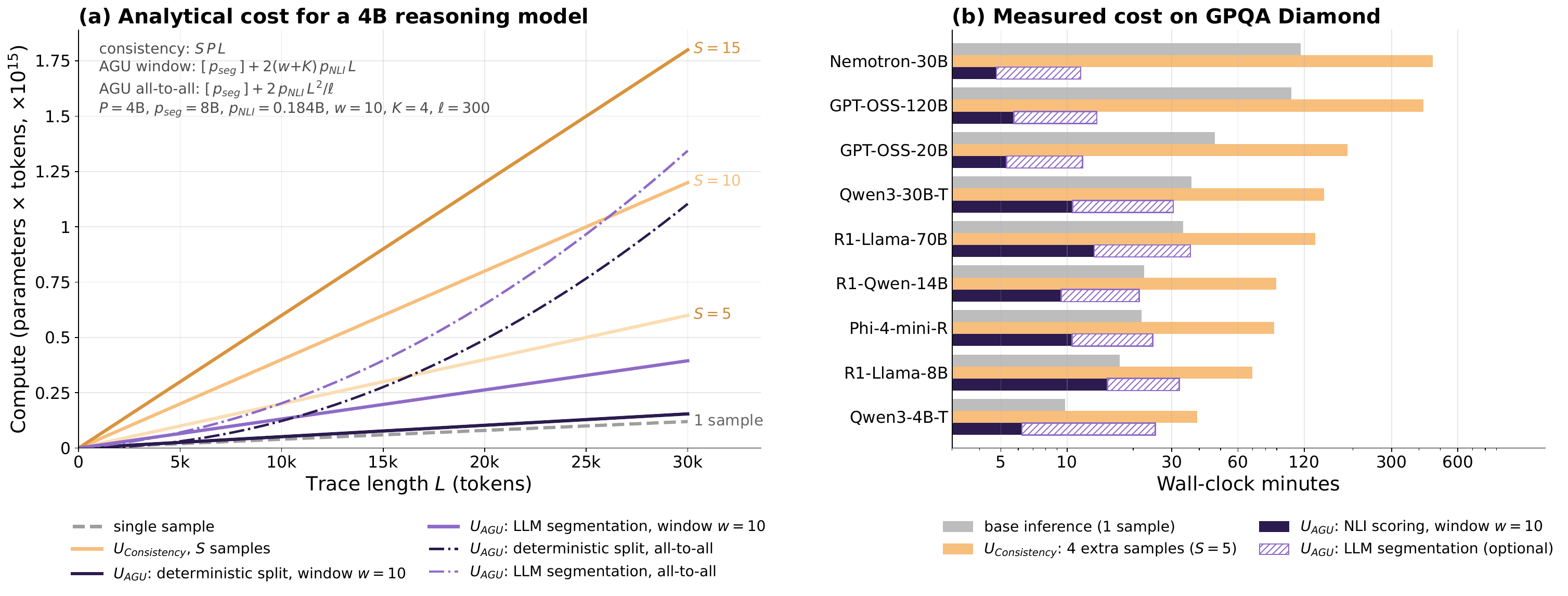}
    \caption{\textbf{Compute cost comparison.}
    (a)~Analytical cost against trace length for the consistency  and  AGU configurations. (b)~Measured wall-clock time per model on GPQA Diamond.}
    \label{fig:computational_costs}
\end{figure}

\section{Discussion}
\label{sec:discussion}

We introduced Argument Graph Uncertainty (AGU), a post-hoc uncertainty
quantification framework that estimates model uncertainty from the logical
structure of a single chain-of-thought reasoning trace. AGU outperforms both
token-level and consistency-based baselines across eight of nine reasoning
models on GPQA Diamond, while remaining computationally cheap. Beyond its
standalone performance, we find that $U_{\mathrm{AGU}}$ captures an orthogonal
uncertainty signal to $U_{\mathrm{Cons}}$: a simple equal-weight blend uniformly
improves over either method alone on GPQA Diamond, MMLU-Pro and TruthfulQA,
even where $U_{\mathrm{Cons}}$ is the stronger single signal. This suggests that internal logical coherence and cross-sample output variance probe
complementary failure modes. Our ablations show that the framework is
robust to its auxiliary components: a deterministic paragraph split matches the
LLM segmenter in mean PRR at zero segmentation cost, the windowed segment graph
matches the all-to-all graph while reducing the NLI stage from quadratic to
linear in the number of segments. We further show that  $U_{\mathrm{AGU}}$ degrades gracefully
under NLI errors, suggesting a robust signal chain. Several directions remain open for future work. First, extending AGU beyond
the MCQA setting, for instance to open-ended generation, requires rethinking
the node-to-option commitment graph, which currently relies on a fixed answer
set. Second, while AGU tolerates edge-level error, the quality of its estimates
remains upstream-dependent on the NLI scorer; a scorer trained on reasoning
text rather than general-domain NLI could yield further calibration gains.

\section{Limitation}
\label{sec:limitation}

AGU depends on the quality of both the segmentation model and the NLI scorer.
Although we show that a deterministic paragraph split suffices and that the
aggregate tolerates substantial edge-level error, NLI models trained on
general-domain corpora can misclassify relations in specialised scientific
language, in particular between short numeric fragments, and our robustness
analysis only covers synthetic perturbations of a single scorer. Our evaluation
is limited to the MCQA setting on three benchmarks; whether AGU generalises to
open-ended generation or other reasoning domains remains an open question.
Finally, the windowed graph keeps the NLI stage linear in trace length, but
very long traces still incur a segmentation cost that scales with the trace
when an LLM segmenter is used.

\paragraph{Societal Impact.}\label{sec:impact} Reliable uncertainty quantification is a prerequisite for 
safe deployment of reasoning models in high-stakes domains. By providing 
interpretable, per-node uncertainty attributions at low computational cost, AGU could lower 
the barrier to auditing model reasoning in such settings. We note, however, that improved 
uncertainty estimates could also increase trust in model outputs in contexts where human 
oversight should be maintained. While we principally see the impact of research in UQ as societally beneficial, we want 
to emphasize that with the rising deployment of agentic systems, well-calibrated 
uncertainty estimates should complement rather than replace human oversight.

\section*{Acknowledgements}
This work was supported by the Senate of Berlin and the European Commision's Digital Europe Programme (DIGITAL) as grant TEF-Health (101100700).

\small
\bibliography{references}


\appendix
\section{Definitions of Metrics and Uncertainty Scores}
\label{appendix:a_experimental_settings}
\subsection{Metrics}

\subsubsection{Prediction Rejection Ratio (PRR).}
PRR measures how well an uncertainty score $u$ can be used to selectively abstain from answering.
Given $N$ samples with correctness indicators $\mathbf{c} = (c_1, \dots, c_N)$ and uncertainty scores for different methods $\mathbf{U(Method)} = (u_1, \dots, u_N)$, we define a \emph{rejection curve} as follows.
For each rejection fraction $r \in [0, 1)$, the $\lfloor rN \rfloor$ samples with the \emph{highest} uncertainty are discarded and accuracy is computed on the remaining $N - \lfloor rN \rfloor$ samples:
\begin{equation}
    \mathrm{Acc}(r;\,\pi) = \frac{1}{N - \lfloor rN \rfloor}
        \sum_{i \in \mathcal{K}(r,\pi)} c_i,
\end{equation}
where $\pi$ is an ordering of samples and $\mathcal{K}(r,\pi)$ denotes the indices of retained samples after rejecting the first $\lfloor rN \rfloor$ elements of $\pi$.
Three orderings are considered:
\begin{itemize}
    \item \textbf{Uncertainty} ($\pi_u$): sort by $u_i$ descending (most uncertain first).
    \item \textbf{Oracle} ($\pi_o$): sort by $c_i$ ascending (incorrect answers first); this is the best achievable ordering.
    \item \textbf{Random} ($\pi_{\mathrm{rand}}$): uniformly random ordering, averaged over $T=200$ independent trials.
\end{itemize}
The area under each rejection curve is
\begin{equation}
    \mathrm{AUC}(\pi) = \int_0^1 \mathrm{Acc}(r;\,\pi)\,\mathrm{d}r,
\end{equation}
approximated numerically with step size $\Delta r = 0.05$.
PRR normalises the uncertainty AUC relative to the oracle and random baselines:
\begin{equation}
    \mathrm{PRR} = \frac{\mathrm{AUC}(\pi_u) - \mathrm{AUC}(\pi_{\mathrm{rand}})}
                        {\mathrm{AUC}(\pi_o)  - \mathrm{AUC}(\pi_{\mathrm{rand}})}.
\end{equation}
$\mathrm{PRR} = 1$ corresponds to perfect oracle rejection; $\mathrm{PRR} = 0$ corresponds to random rejection; negative values indicate that the uncertainty score is anti-correlated with error.

\begin{figure}[!h]
  \centering
  \includegraphics[width=0.5\linewidth]{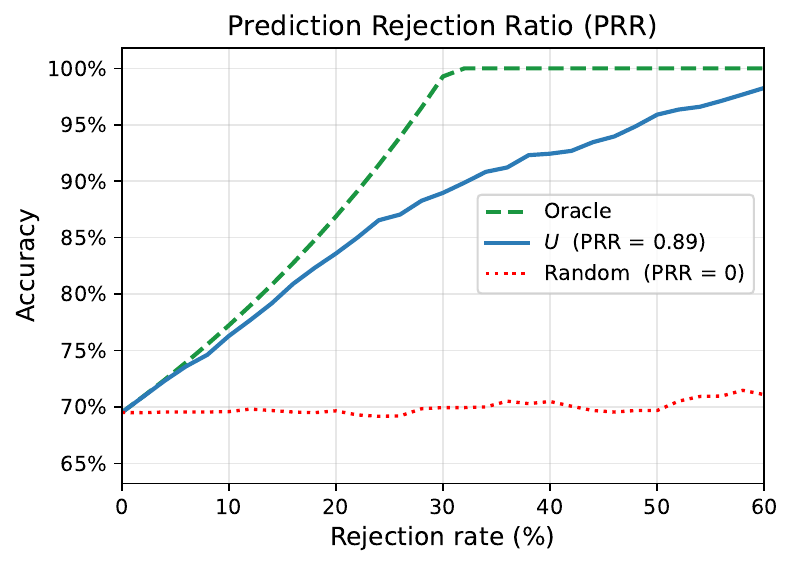}
  \caption{Example visualization of PRR.}
\end{figure}

\subsubsection{Spearman Rank Correlation.}
To measure the monotonic relationship between an uncertainty score $u_i \in \mathbb{R}$ and binary correctness $c_i \in \{0, 1\}$ for a set of $N$ questions, we compute the Spearman rank correlation
\begin{equation}
    \rho = \frac{\sum_{i=1}^{N}(r_i - \bar{r})(s_i - \bar{s})}
               {\sqrt{\sum_{i=1}^{N}(r_i - \bar{r})^2 \; \sum_{i=1}^{N}(s_i - \bar{s})^2}},
\end{equation}
where $r_i = \operatorname{rank}(u_i)$ and $s_i = \operatorname{rank}(c_i)$ are the average-tied ranks of the uncertainty scores and correctness labels, respectively, and $\bar{r}, \bar{s}$ denote their means.
A negative $\rho$ indicates that higher uncertainty is associated with lower correctness, which is the desired behaviour for a well-calibrated uncertainty estimate.

\subsection{Uncertainty Scores}

\subsubsection{ Argument Graph Uncertainty.}
$U_{\mathrm{AGU}} = H_{\mathrm{norm}}$ as defined in
Section~\ref{sec:method}.

\subsubsection{Consistency Uncertainty.}
$U_{\mathrm{cons}}$ measures disagreement across $S$ independent runs of the same
model on the same question. Let $c_k$ be the number of runs that predicted option
$k$, and $\hat{p}_k = c_k / S$ the empirical vote fraction. The consistency
uncertainty is the normalised Shannon entropy of the vote distribution:
\begin{equation}
    U_{\mathrm{cons}} =
    \frac{-\sum_{k=1}^{K} \hat{p}_k \ln \hat{p}_k}{\ln K},
\end{equation}
where the sum is taken only over options that received at least one vote. A value
of $0$ indicates unanimous agreement across runs; a value of $1$ indicates
maximally uniform disagreement.

\subsubsection{Token-Level Uncertainty.}
$U_{\mathrm{token}}$ is derived from the log-probabilities the model assigns to
each answer-option letter at the final answer token position. Let
$\ell_k = \log p_\theta(\text{option } k \mid \text{context})$ be the log-probability
of option $k$ extracted from the model's top-$\kappa$ logits. These are converted
to a probability distribution via a softmax over the $K$ options:
\begin{equation}
    p_k = \frac{\exp(\ell_k)}{\sum_{j=1}^{K} \exp(\ell_j)}.
\end{equation}
The token uncertainty is then the normalised Shannon entropy of this distribution:
\begin{equation}
    U_{\mathrm{token}} =
    \frac{-\sum_{k=1}^{K} p_k \ln p_k}{\ln K}.
\end{equation}

\section{Additional Results}
\label{sec:appendix_b_additional_results}

\subsection{Additional Benchmarks}
\label{appendix:additional_benchmarks}

Tables~\ref{tab:mmlu_pro} and~\ref{tab:truthfulqa} report PRR on MMLU-Pro and
TruthfulQA MC4 for the same signals as \ref{tab:main_table}. Among the single signals, $U_{\mathrm{AGU}}$
is the best on two of five MMLU-Pro models and two of nine TruthfulQA models,
with $U_{\mathrm{Cons}}$ leading elsewhere and $U_{\mathrm{Token}}$ only once.
The equal-weight combination $U_{\mathrm{AGU}}{+}U_{\mathrm{Cons}}$ is the best
signal for every model on both benchmarks, improving on the stronger of its two
components by $0.10$ PRR on average on MMLU-Pro and $0.09$ on TruthfulQA. This further empirically supports our finding that $U_{\mathrm{AGU}}$ is orthogonal to
consistency-based uncertainty.

\begin{table}[!h]
\centering
\footnotesize
\setlength{\tabcolsep}{5pt}
\caption{\textbf{PRR on MMLU-Pro.} Mean $\pm$ std over seeds. Bold marks the best signal per model.}
\label{tab:mmlu_pro}
\begin{tabular}{lcccc}
\toprule
\textbf{Model} & $U_{\mathrm{AGU}}$ & $U_{\mathrm{Token}}$ & $U_{\mathrm{Cons}}$ & $U_{\mathrm{AGU}}{+}U_{\mathrm{Cons}}$ \\
\midrule
GPT-OSS-120B & 0.54 $\pm$ .01 & 0.05 $\pm$ .00 & 0.48 $\pm$ .01 & \textbf{0.66 $\pm$ .01} \\
Nemotron-30B & 0.52 $\pm$ .01 & 0.11 $\pm$ .01 & 0.51 $\pm$ .00 & \textbf{0.67 $\pm$ .00} \\
Qwen3-30B-T  & 0.43 $\pm$ .01 & 0.05 $\pm$ .02 & 0.49 $\pm$ .00 & \textbf{0.63 $\pm$ .01} \\
Phi-4-mini-R & 0.41 $\pm$ .01 & 0.07 $\pm$ .00 & 0.60 $\pm$ .01 & \textbf{0.64 $\pm$ .00} \\
R1-Llama-70B & 0.40 $\pm$ .01 & 0.04 $\pm$ .01 & 0.51 $\pm$ .00 & \textbf{0.63 $\pm$ .01} \\
\midrule
Mean         & 0.46 & 0.06 & 0.52 & \textbf{0.64} \\
\bottomrule
\end{tabular}
\end{table}

\begin{table}[!h]
\centering
\footnotesize
\setlength{\tabcolsep}{5pt}
\caption{\textbf{PRR on TruthfulQA MC4.} Mean $\pm$ std over seeds. Bold marks the best signal per model.}
\label{tab:truthfulqa}
\begin{tabular}{lcccc}
\toprule
\textbf{Model} & $U_{\mathrm{AGU}}$ & $U_{\mathrm{Token}}$ & $U_{\mathrm{Cons}}$ & $U_{\mathrm{AGU}}{+}U_{\mathrm{Cons}}$ \\
\midrule
Qwen3-30B-T  & 0.48 $\pm$ .05 & 0.13 $\pm$ .02 & 0.37 $\pm$ .03 & \textbf{0.60 $\pm$ .03} \\
R1-Llama-70B & 0.42 $\pm$ .04 & 0.33 $\pm$ .01 & 0.44 $\pm$ .02 & \textbf{0.59 $\pm$ .02} \\
Qwen3-4B-T   & 0.39 $\pm$ .04 & 0.18 $\pm$ .04 & 0.36 $\pm$ .03 & \textbf{0.51 $\pm$ .04} \\
Phi-4-mini-R & 0.38 $\pm$ .03 & 0.11 $\pm$ .05 & 0.46 $\pm$ .02 & \textbf{0.55 $\pm$ .02} \\
GPT-OSS-120B & 0.35 $\pm$ .03 & 0.11 $\pm$ .02 & 0.37 $\pm$ .03 & \textbf{0.48 $\pm$ .03} \\
Nemotron-30B & 0.34 $\pm$ .04 & 0.17 $\pm$ .03 & 0.46 $\pm$ .01 & \textbf{0.51 $\pm$ .01} \\
R1-Qwen-14B  & 0.34 $\pm$ .03 & 0.49 $\pm$ .03 & 0.47 $\pm$ .01 & \textbf{0.55 $\pm$ .02} \\
GPT-OSS-20B  & 0.29 $\pm$ .06 & 0.11 $\pm$ .03 & 0.45 $\pm$ .03 & \textbf{0.53 $\pm$ .04} \\
R1-Llama-8B  & 0.24 $\pm$ .05 & 0.25 $\pm$ .02 & 0.33 $\pm$ .02 & \textbf{0.41 $\pm$ .05} \\
\midrule
Mean         & 0.36 & 0.21 & 0.41 & \textbf{0.53} \\
\bottomrule
\end{tabular}
\end{table}


\subsection{NLI Edge Statistics}
\label{appendix:edge_stats}

\begin{figure*}[!t]
  \centering
  \includegraphics[width=\textwidth]{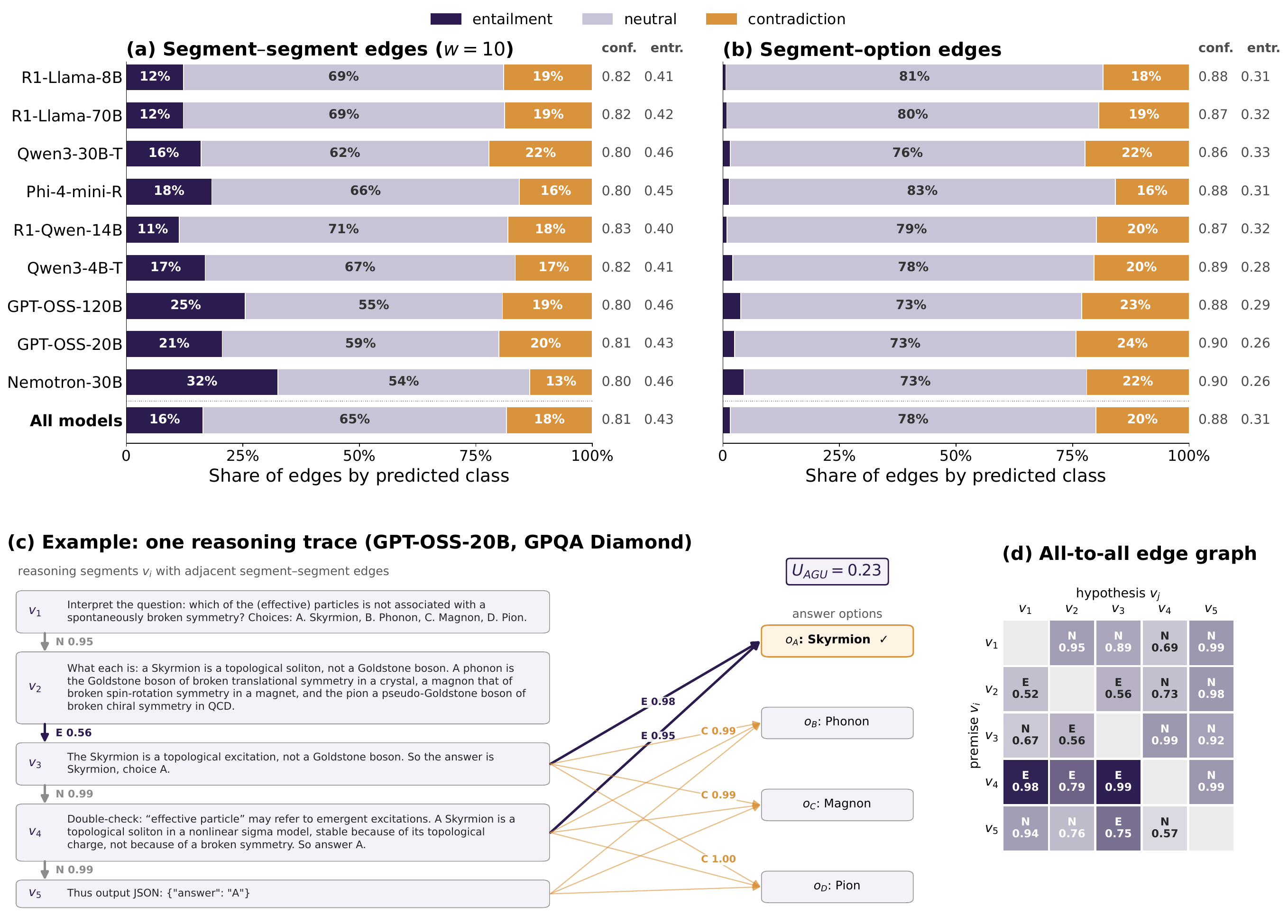}
  \caption{\textbf{NLI edge statistics.}
  (a,~b)~Share of edges by predicted class per model for segment--segment edges
  within the window ($w{=}10$) and segment--option edges; conf.\ and entr.\ are
  the mean probability of the predicted class and the mean three-class entropy
  normalised by $\log 3$.
  (c)~A complete five-segment trace of GPT-OSS-20B with the scorer's actual
  outputs; segment text is abbreviated. Adjacent segment--segment edges are shown
  along the chain. For visibility, only segment--option edges with $p \geq 0.75$
  are drawn as arrows; $v_1$ and $v_2$ are neutral towards all options and $v_5$
  towards $o_A$.
  (d)~All ordered segment--segment edges of the same trace; row is premise,
  column is hypothesis, shading is the scorer's confidence.}
  \label{fig:edge_stats}
\end{figure*}

Fig.~\ref{fig:edge_stats} provides summary statistics for GPQA diamond. Neutral dominates both edge types (65\% of
segment--segment, 78\% of segment--option edges), as consecutive reasoning
steps mostly add new content and most steps neither support nor rule out a
given option. Segment--option edges are almost never entailment (1.6\%) but
contradiction in 20\% of cases, so the per-option evidence rests on a few
confident entailment edges from the concluding segments, while contradiction
edges rule out the remaining options. The profile is generally stable across models; the segment--segment entailment share can varies notably (11\% to 32\%). The scorer is more decisive on segment--option edges
(confidence 0.88 vs.\ 0.81, entropy 0.31 vs.\ 0.43); this is the channel that enters the
Dirichlet evidence directly. The example in panels~(c,~d) shows a correctly
answered trace with $U_{\mathrm{AGU}} = 0.23$: the two concluding segments
entail the correct option at $p \geq 0.95$ and contradict the others at
$p \geq 0.99$, and the double-check segment $v_4$ entails all preceding steps,
giving it the highest coherence weight.

\subsection{Analysis of NLI Robustness}
\label{appendix:nli_robustness}

We measure how errors of the NLI scorer at the level of individual edges
propagate to $U_{\mathrm{AGU}}$. For each of the nine GPQA Diamond models we
take the scorer's class probabilities for every segment--segment and
segment--option edge, perturb them, and recompute $U_{\mathrm{AGU}}$ from the
perturbed edges. Each perturbation is applied at several
noise levels with three random seeds per level, and we report PRR as a
fraction of the PRR obtained from the unperturbed edges, averaged over seeds
and models.

\paragraph{Random label flips.}
With probability $\varepsilon$ per edge, the scorer's predicted class is
replaced by one of the two other classes chosen uniformly at random, for
$\varepsilon$ from $0.05$ to $0.4$. This simulates hard classification errors
that are uncorrelated with the content of the edge. PRR decays smoothly and
stays positive for every model: 90\% of the clean PRR is retained at
$\varepsilon{=}0.05$, 78\% at $0.1$, 62\% at $0.2$ and 52\% when 40\% of all
edges carry a wrong label. The decline flattens beyond $\varepsilon{=}0.15$,
since the remaining signal is carried by the majority of edges that are still
correct.

\paragraph{Probability jitter.}
Gaussian noise $\eta \sim \mathcal{N}(0, \sigma^{2})$ is added to the class
log-probabilities of every edge, which are then renormalised,
$\tilde{p} \propto \exp(\log p + \eta)$, for $\sigma$ from $0.25$ to $2$. This
simulates a miscalibrated scorer: the predicted class changes only where the
margin between classes is smaller than the noise. PRR retention is 97\% at
$\sigma{=}0.5$, 85\% at $\sigma{=}1$ and 65\% at $\sigma{=}2$, where the noise
exceeds the typical margin and a large share of predicted classes change.
$U_{\mathrm{AGU}}$ thus relies on the scorer's ordinal judgment rather than on
calibrated probabilities.

Both quantities that enter the Dirichlet parameters are averages over many
edges, so a single misclassified edge shifts the per-option evidence only
marginally.

\subsection{Windowed vs.\ All-to-All Segment Graph}
\label{appendix:window_ablation}

Scoring every ordered segment pair is quadratic in the number of segments and
could become a bottleneck for long traces. The windowed variant of
Section~\ref{sec:all_to_all} scores only pairs with $|i-j| \le w/2$, so that
with $w=10$ each segment is compared against its five preceding and five
following segments in both directions. All segment--option edges are still
scored, and Conflict, Asym and the global coherence become averages over the
windowed edge set; the local trajectory term is unaffected. We compare the two
graphs on GPQA Diamond for all nine models
(Table~\ref{tab:window_ablation}). Both columns are computed from the same NLI
outputs of a single seed per model, so this is a paired comparison; the
absolute PRR values therefore differ from the five-seed means of
Table~\ref{tab:main_table}, but lie within the reported standard deviations.

The window uses 18 times fewer segment--segment pairs while PRR is essentially
unchanged, 0.42 against 0.42 on average and at most 0.01 apart for any
model. We attribute this to the locality of logical structure in a reasoning
trace: a step supports or attacks the steps in its immediate neighbourhood,
while distant pairs are predominantly neutral and contribute little signal.
The NLI stage can therefore stay linear in the number of segments as traces
grow, without affecting the quality of the uncertainty estimate.

\begin{table}[!h]
\centering
\footnotesize
\setlength{\tabcolsep}{4pt}
\caption{\textbf{Windowed ($w{=}10$) vs.\ all-to-all segment graph on GPQA Diamond.}
PRR and number of scored segment--segment pairs, one seed per model.}
\label{tab:window_ablation}
\begin{tabular}{lrcccc}
\toprule
& & \multicolumn{2}{c}{\textbf{PRR}} & \multicolumn{2}{c}{\textbf{Pairs}} \\
\cmidrule(lr){3-4}\cmidrule(lr){5-6}
\textbf{Model} & \textbf{Mean segs} & \textbf{Window} & \textbf{All-to-all} & \textbf{Window} & \textbf{All-to-all} \\
\midrule
GPT-OSS-20B   &  69.4 & 0.67 & 0.67 &  94.9k & 1.64M \\
GPT-OSS-120B  &  61.8 & 0.59 & 0.60 & 102.3k & 1.43M \\
Nemotron-30B  &  64.4 & 0.54 & 0.53 &  85.3k & 1.02M \\
Qwen3-30B-T   &  99.2 & 0.48 & 0.47 & 190.4k & 3.51M \\
R1-Qwen-14B   &  92.8 & 0.34 & 0.33 & 168.8k & 2.86M \\
Qwen3-4B-T    &  59.7 & 0.33 & 0.33 & 111.2k & 1.48M \\
R1-Llama-70B  & 123.7 & 0.32 & 0.32 & 239.1k & 4.53M \\
R1-Llama-8B   & 146.9 & 0.28 & 0.27 & 274.9k & 6.88M \\
Phi-4-mini-R  &  98.6 & 0.28 & 0.27 & 189.3k & 3.45M \\
\midrule
Mean / total  &  90.7 & 0.42 & 0.42 & 1.46M  & 26.81M \\
\bottomrule
\end{tabular}
\end{table}

\subsection{Trace Length versus Graph Signal}
\label{appendix:length}

We test whether $U_{\mathrm{AGU}}$ carries information beyond the length of the
reasoning trace. For each GPQA Diamond model we compute three
uncertainty scores on the same items and evaluate each by PRR: (i)~the number of
segments $n$ used directly as the score, (ii)~$U_{\mathrm{AGU}}$, and
(iii)~$U_{\mathrm{AGU}}$ with the length component removed. For (iii) we
convert both $U_{\mathrm{AGU}}$ and $n$ to ranks, fit a linear regression of
$\mathrm{rank}(U_{\mathrm{AGU}})$ on $\mathrm{rank}(n)$, and use the residuals
as the score; this removes any monotone dependence on length while keeping the
remaining ordering of $U_{\mathrm{AGU}}$. We also report the Spearman
correlation $\rho$ between $U_{\mathrm{AGU}}$ and $n$.

Segment count alone attains a mean PRR of $0.33$ against $0.45$ for
$U_{\mathrm{AGU}}$, and is the weaker signal on seven of nine models. The two
scores are moderately correlated ($\rho = 0.43$ on average). Some correlation
is expected, since harder questions elicit both longer traces and more
internal conflict, so length and graph structure partly respond to the same
underlying difficulty. Correlation does not imply redundancy, however: the
residual score retains a PRR of $0.33$, so the majority of the graph signal is
independent of length. The residual is smallest for GPT-OSS-120B and
R1-Llama-70B, the two models with the longest traces, where length accounts for
the largest share of the signal.

\begin{table}[!h]
\centering
\footnotesize
\setlength{\tabcolsep}{4pt}
\caption{\textbf{Trace length versus graph signal on GPQA Diamond.} PRR of the
segment count used as an uncertainty score, of $U_{\mathrm{AGU}}$, and of
$U_{\mathrm{AGU}}$ after rank-regressing out the segment count; $\rho$ is the
Spearman correlation between $U_{\mathrm{AGU}}$ and segment count. Mean $\pm$
std over seeds; bold marks the better of the two single scores.}
\label{tab:length}
\begin{tabular}{lcccc}
\toprule
\textbf{Model} & \textbf{Segment count} & $U_{\mathrm{AGU}}$ & $U_{\mathrm{AGU}}\,|\,$length & $\rho$ \\
\midrule
GPT-OSS-20B  & 0.52 $\pm$ .06 & \textbf{0.67 $\pm$ .06} & 0.49 $\pm$ .08 & 0.45 $\pm$ .07 \\
Nemotron-30B & 0.46 $\pm$ .14 & \textbf{0.60 $\pm$ .08} & 0.45 $\pm$ .11 & 0.52 $\pm$ .07 \\
GPT-OSS-120B & \textbf{0.65 $\pm$ .03} & 0.58 $\pm$ .02 & 0.28 $\pm$ .02 & 0.59 $\pm$ .01 \\
Qwen3-30B-T  & 0.41 $\pm$ .07 & \textbf{0.50 $\pm$ .05} & 0.38 $\pm$ .05 & 0.54 $\pm$ .04 \\
R1-Qwen-14B  & 0.23 $\pm$ .09 & \textbf{0.37 $\pm$ .04} & 0.31 $\pm$ .06 & 0.29 $\pm$ .06 \\
Qwen3-4B-T   & 0.15 $\pm$ .04 & \textbf{0.35 $\pm$ .02} & 0.29 $\pm$ .05 & 0.36 $\pm$ .07 \\
R1-Llama-70B & \textbf{0.34 $\pm$ .03} & 0.33 $\pm$ .04 & 0.22 $\pm$ .05 & 0.46 $\pm$ .04 \\
Phi-4-mini-R & 0.06 $\pm$ .07 & \textbf{0.32 $\pm$ .06} & 0.32 $\pm$ .08 & 0.29 $\pm$ .07 \\
R1-Llama-8B  & 0.13 $\pm$ .06 & \textbf{0.28 $\pm$ .04} & 0.22 $\pm$ .04 & 0.31 $\pm$ .09 \\
\midrule
Mean         & 0.33 & \textbf{0.45} & 0.33 & 0.43 \\
\bottomrule
\end{tabular}
\end{table}

\subsection{Ablation on Uncertainty Score Combination}
\label{combination_ablation}
\begin{figure}[!h]
  \centering
  \label{fig:boxplot_combos}
  \includegraphics[width=\linewidth]{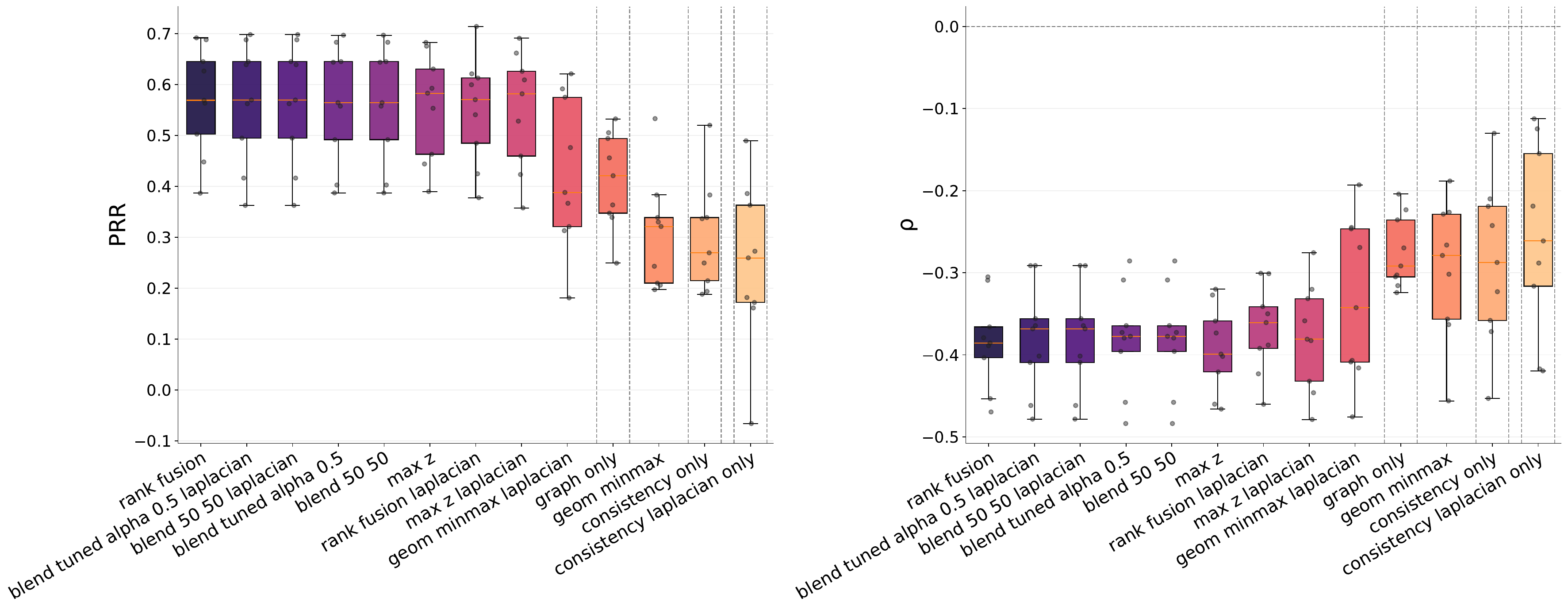}
  \caption{Different uncertainty score combination techniques aggregated over models.}
\end{figure}

In this section we provide additional experimental results. 
 $u_{\mathrm{agu}}$ denotes graph-based uncertainty and $u_{\mathrm{cons}}$ denotes a consistency uncertainty signal
(either normalized vote entropy or Laplacian-based consistency uncertainty).
For each model, combinations are computed at the question level.

\paragraph{Normalization primitives.}
Given vectors of per-question uncertainties, we define:
\begin{align}
z_{\mathrm{agu}} &= \frac{u_{\mathrm{agu}} - \mu_{\mathrm{agu}}}{\sigma_{\mathrm{agu}}}, &
z_{\mathrm{cons}} &= \frac{u_{\mathrm{cons}} - \mu_{\mathrm{cons}}}{\sigma_{\mathrm{cons}}}, \\
m_{\mathrm{agu}} &= \frac{u_{\mathrm{agu}} - \min(u_{\mathrm{agu}})}{\max(u_{\mathrm{agu}})-\min(u_{\mathrm{agu}})}, &
m_{\mathrm{cons}} &= \frac{u_{\mathrm{cons}} - \min(u_{\mathrm{cons}})}{\max(u_{\mathrm{cons}})-\min(u_{\mathrm{cons}})}, \\
r_{\mathrm{agu}} &= \mathrm{rankpct}(u_{\mathrm{agu}}), &
r_{\mathrm{cons}} &= \mathrm{rankpct}(u_{\mathrm{cons}}),
\end{align}
where $\mathrm{rankpct}(\cdot)$ is the percentile rank in $[0,1]$.
If a denominator is zero (constant signal), the corresponding normalized values are set to $0$.

\paragraph{Laplacian consistency uncertainty.}
Given $m$ sampled predictions (votes) $\{v_i\}_{i=1}^m$, define the agreement matrix
\[
W_{ij}=\mathbf{1}[v_i=v_j].
\]
Let $d_i=\sum_{j=1}^m W_{ij}$ and $D=\mathrm{diag}(d_1,\ldots,d_m)$.  
The normalized graph Laplacian is
\[
L = I - D^{-1/2} W D^{-1/2}.
\]
If $\{\lambda_k\}_{k=1}^m$ are eigenvalues of $L$, the Laplacian consistency uncertainty is
\[
u_{\mathrm{cons}}^{\mathrm{lap}} = \sum_{k=1}^m \max(0,\,1-\lambda_k).
\]

\paragraph{Blending methods.}
We evaluate the following fused uncertainty scores:
\begin{align}
\text{Equal-weight z-blend (blend\_50\_50):}\quad
s_{\mathrm{blend50}} &= \tfrac{1}{2}z_{\mathrm{agu}} + \tfrac{1}{2}z_{\mathrm{cons}}, \\[2mm]
\text{Tuned-$\alpha$ z-blend (blend\_tuned\_alpha\_$\alpha$):}\quad
s_{\mathrm{blend},\alpha} &= \alpha z_{\mathrm{agu}} + (1-\alpha)z_{\mathrm{cons}},\quad \alpha\in[0,1], \\[2mm]
\text{Rank fusion (rank\_fusion):}\quad
s_{\mathrm{rank}} &= \tfrac{1}{2}r_{\mathrm{agu}} + \tfrac{1}{2}r_{\mathrm{cons}}, \\[2mm]
\text{Max-z fusion (max\_z):}\quad
s_{\max z} &= \max(z_{\mathrm{agu}}, z_{\mathrm{cons}}), \\[2mm]
\text{Geometric min--max fusion (geom\_minmax):}\quad
s_{\mathrm{geom}} &= \sqrt{m_{\mathrm{agu}}\,m_{\mathrm{cons}}}.
\end{align}

\paragraph{Tuning of $\alpha$.}
For the tuned blend, we search $\alpha\in\{0.0,0.1,\dots,1.0\}$ and select
\begin{equation}
\alpha^\star = \arg\max_{\alpha}\ \frac{1}{|\mathcal{M}|}\sum_{m\in\mathcal{M}} \mathrm{PRR}_m(\alpha),
\end{equation}
where $\mathrm{PRR}_m(\alpha)$ is the PRR of model $m$ using $s_{\mathrm{blend},\alpha}$.

\paragraph{Consistency variants.}
All formulas above are applied with:
\begin{itemize}
    \item $u_{\mathrm{cons}} = u_{\mathrm{cons}}$ (as defined in \ref{appendix:a_experimental_settings}), and
    \item $u_{\mathrm{cons}} = u_{\mathrm{cons}}^{\mathrm{lap}}$,
\end{itemize}
yielding the corresponding ``\_laplacian'' method variants in the latter case in Fig.\ref{fig:boxplot_combos}.

\subsection{Uncertainty Score Distributions}

\begin{figure}[!h]
  \centering
  \includegraphics[width=\linewidth]{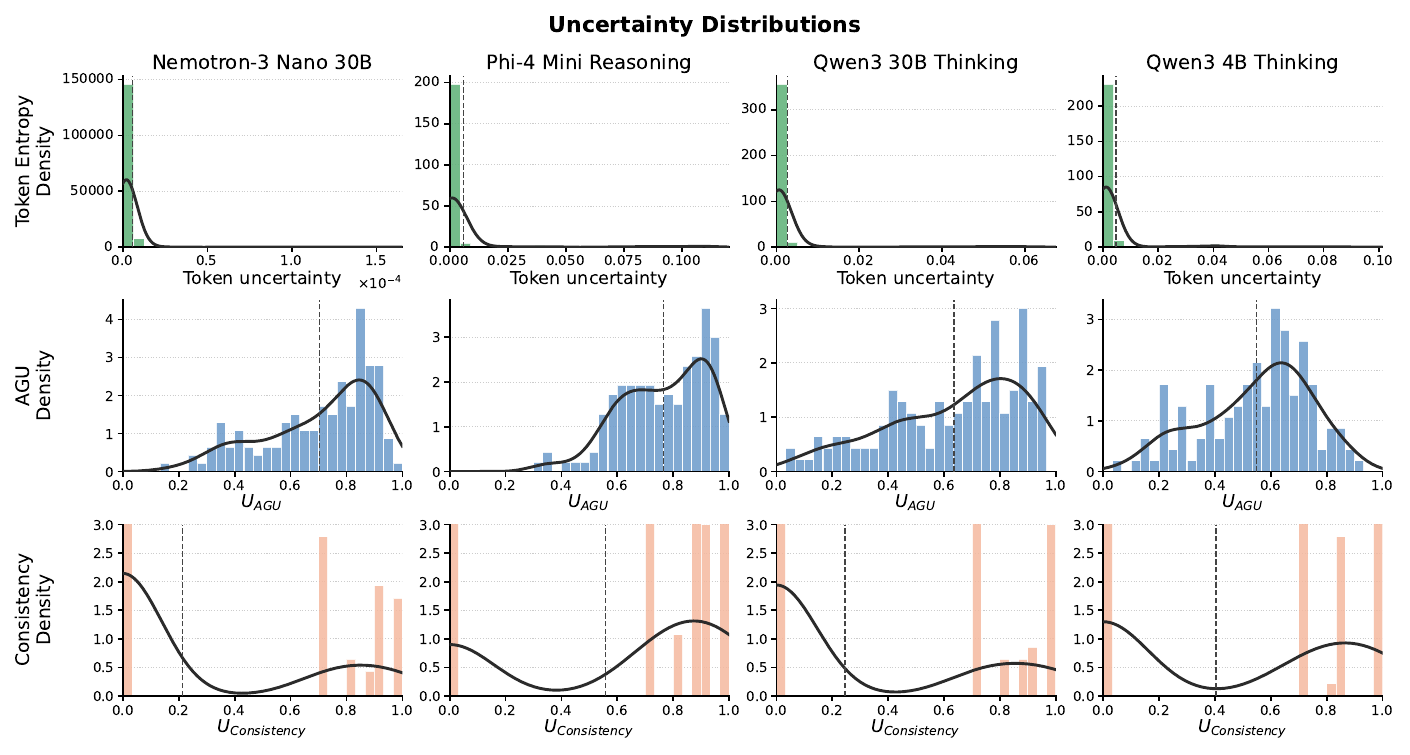}
  \caption{Histograms of uncertainty distributions per model and method.}
  \label{fig:histograms_2}
\end{figure}
Figure \ref{fig:histograms_2} offers uncertainty scores distributions for the models not shown in the main manuscript. Distributions are consistent across all models.

\section{Computational Costs}
\label{appedix:computational_costs}
\paragraph{Notation.}
Let $P$ denote the parameter count of the evaluated reasoning model,
$L_p$ the prompt length and $L$ the length of one reasoning trace $T$ in tokens,
and $s$ the number of sampled traces used by the consistency baseline
($s=5$ in our experiments). We measure compute in parameter--token products,
which is proportional to FLOPs for a transformer forward pass
(${\approx}\,2P$ FLOPs per token).

\paragraph{Consistency-based uncertainty.}
Every sample requires a full forward pass of the evaluated model over the
prompt and an autoregressive decode of a fresh trace $T$,
\begin{equation}
  C_{\mathrm{sample}} \;=\; P\,(L_p + L) \;\in\; \mathcal{O}(P\,L),
  \qquad L_p \ll L .
\end{equation}
Since one trace is generated in any case to obtain the answer,
the \emph{overhead} of $U_{\mathrm{Cons}}$ is the cost of the $s-1$ additional traces,
\begin{equation}
  C_{\mathrm{Cons}} \;=\; (s-1)\,P\,(L_p + L) \;\in\; \mathcal{O}(s\,P\,L).
  \label{eq:cost_cons}
\end{equation}
The overhead therefore scales linearly with the size of the evaluated model and
with trace length, and is incurred on the evaluated model itself, i.e.\ at its
full GPU footprint. 

\paragraph{Argument Graph Uncertainty.}
AGU operates post hoc on the single trace $T$ that is generated anyway and makes no
further calls to the evaluated model. Its overhead is incurred entirely on two
small auxiliary models: a segmenter with $p_{\mathrm{seg}}$ parameters and an NLI
encoder with $p_{\mathrm{NLI}}$ parameters. Let $T$ be split into $n$
segments of average length $\ell = L/n$ tokens, and let $K$ denote the number of
answer options.

\emph{Segmentation.}
With the instruction-tuned segmenter, $T$ is read once as a prompt and a
short list of segment boundaries is emitted,
\begin{equation}
  C_{\mathrm{seg}} \;=\; p_{\mathrm{seg}}\,(L + \mathcal{O}(n)) \;\in\; \mathcal{O}(p_{\mathrm{seg}}\,L).
\end{equation}
With the deterministic paragraph split this stage requires no model call and
$C_{\mathrm{seg}} = 0$.

\emph{NLI scoring.}
Every scored pair is one encoder pass over a premise and a hypothesis of
${\approx}\,2\ell$ tokens in total, so the cost is the number of pairs times
$2\ell\,p_{\mathrm{NLI}}$. The segment--option graph always contributes $nK$
pairs. The segment--segment graph contributes $n(n-1)$ pairs for the all-to-all
variant and ${\approx}\,n\,w$ pairs for a window of width $w$:
\begin{align}
  C_{\mathrm{NLI}}^{\mathrm{window}}
    &= \big(n\,w + n\,K\big)\,2\ell\,p_{\mathrm{NLI}}
     = 2\,(w + K)\,p_{\mathrm{NLI}}\,L
     \;\in\; \mathcal{O}\!\big(p_{\mathrm{NLI}}\,(w{+}K)\,L\big),
  \label{eq:cost_nli_window}\\[2pt]
  C_{\mathrm{NLI}}^{\mathrm{all}}
    &= \big(n(n-1) + n\,K\big)\,2\ell\,p_{\mathrm{NLI}}
     \;\approx\; 2\,p_{\mathrm{NLI}}\,\frac{L^{2}}{\ell}
     \;\in\; \mathcal{O}\!\big(p_{\mathrm{NLI}}\,L^{2}/\ell\big).
  \label{eq:cost_nli_all}
\end{align}
The windowed cost is linear in $L$ and independent of $n$: a finer segmentation
produces more pairs, but each pair is proportionally shorter. Only the all-to-all
graph is quadratic in trace length.

\emph{Total overhead and comparison.}
Combining the two stages gives four configurations,
\begin{equation}
  C_{\mathrm{AGU}} \;=\; \underbrace{[\,p_{\mathrm{seg}}\,L\,]}_{\text{LLM segmenter only}}
  \;+\;
  \begin{cases}
    2\,(w+K)\,p_{\mathrm{NLI}}\,L & \text{window } w,\\[2pt]
    2\,p_{\mathrm{NLI}}\,L^{2}/\ell & \text{all-to-all}.
  \end{cases}
  \label{eq:cost_agu}
\end{equation}
For the windowed graph, the ratio to the
consistency overhead of Eq.~\eqref{eq:cost_cons} is independent of trace length,
\begin{equation}
  \frac{C_{\mathrm{AGU}}}{C_{\mathrm{Cons}}}
  \;=\;
  \frac{[\,p_{\mathrm{seg}}\,] + 2\,(w+K)\,p_{\mathrm{NLI}}}{(s-1)\,P}
  \;\in\; \mathcal{O}\!\left(\frac{p_{\mathrm{NLI}}}{P}\right),
  \label{eq:cost_ratio}
\end{equation}
and decreases as the evaluated model grows.

\subsection{Hyper Parameter Ablation}
\label{appendix_hyper_params}

We conduct a sweep over $\kappa$ and $\alpha_0$ with $\gamma_c$ and $\gamma_a$ fixed at 
their default values of $(1.2, 1.0)$, selecting the combination that maximises PRR per 
model. Table~\ref{tab:hyper_params} reports 
the best-found configuration alongside, with bold entries indicating 
models that improve over the default $(\kappa=12.0, \alpha_0=1.0)$ setting.

Five of nine models already achieve their best performance at the default 
hyperparameters. The remaining four models show moderate gains 
under tuned configurations, typically favouring a larger $\kappa$ and smaller $\alpha_0$, 
which concentrate the Dirichlet evidence more aggressively and reduce the influence of 
the symmetric prior.

\begin{figure}[!h]
  \centering
  \includegraphics[width=\linewidth]{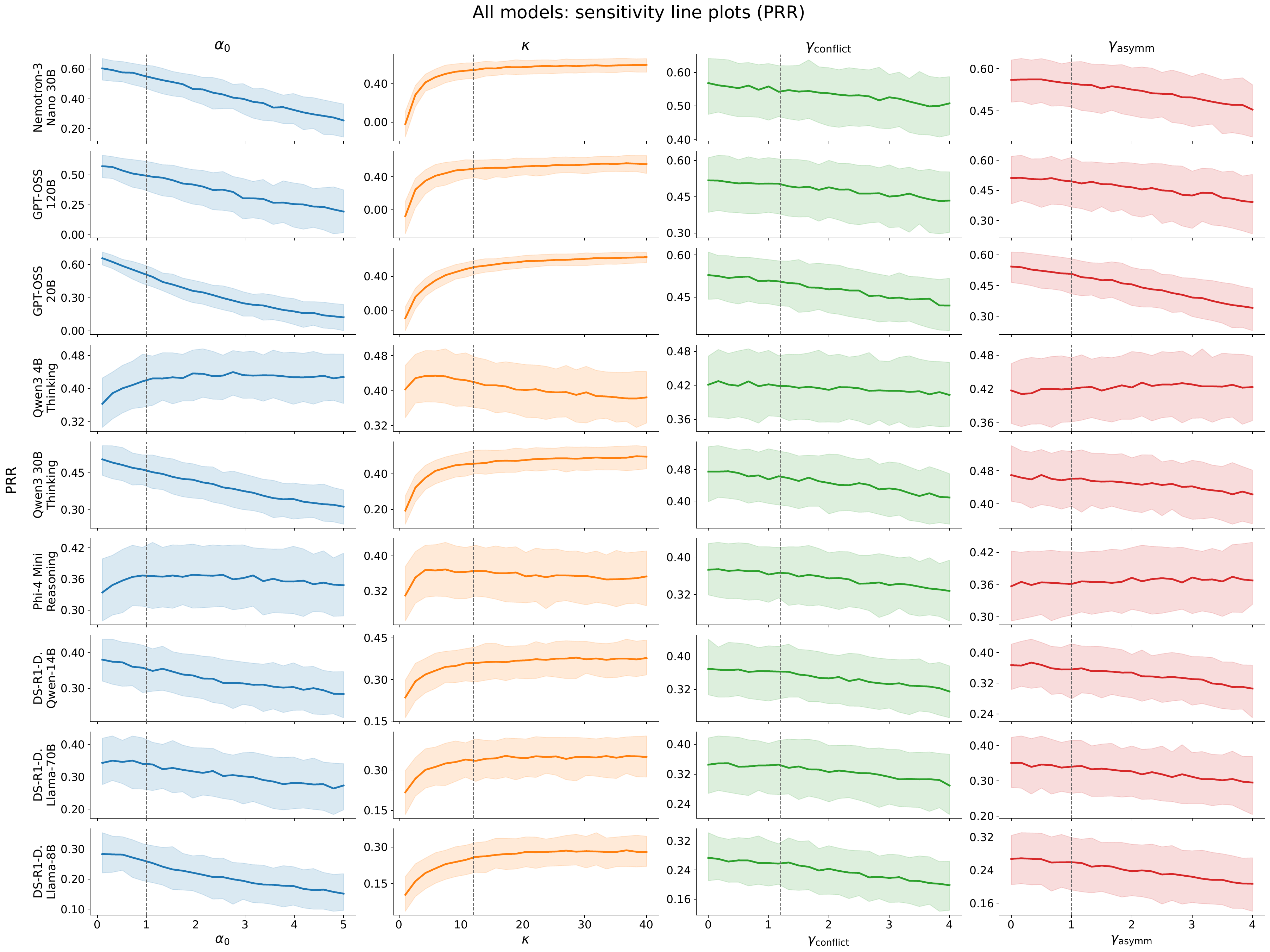}
  \caption{Influence of hyperparameters evaluated for every model.}
  \label{fig:hyperparams_lineplots}
\end{figure}

The sensitivity plots in Figure~\ref{fig:hyperparams_lineplots} shows 
PRR is largely stable across a broad range of $\kappa$ and $\alpha_0$ values, with 
performance degrading only at extreme settings. $\gamma_{\mathrm{conflict}}$ and 
$\gamma_{\mathrm{asymm}}$ exhibit similarly flat sensitivity curves, indicating that 
the penalty terms in Equation~\ref{eq:kappa_eff} are not critical to tune precisely. 
Taken together, these results support the use of a single fixed hyperparameter 
configuration across all models, as adopted in the main experiments, while careful choice of hyperparameters can lead to moderate improvments in some cases.

\begin{table}[!h]
\centering
\footnotesize
\setlength{\tabcolsep}{4pt}
\caption{Results of sweep over $\kappa,\alpha_0$ and fixed $(\gamma_c,\gamma_a)$. Improved models are \textbf{bold}.}
\label{tab:hyper_params}
\begin{tabular}{lcccc}
\toprule
\textbf{Model} & \textbf{PRR $\uparrow$} & \textbf{$\rho$ $\downarrow$} & \textbf{$(\kappa,\alpha_0)$} & \textbf{$(\gamma_c,\gamma_a)$} \\
\midrule
\textbf{GPT-OSS-20B}  & \textbf{0.66} & \textbf{-0.39} & \textbf{(27.68, 0.10)} & \textbf{(1.20, 1.00)} \\
Nemotron-30B          & 0.65 & -0.37 & (12.00, 1.00) & (1.20, 1.00) \\
\textbf{GPT-OSS-120B} & \textbf{0.68} & \textbf{0.26}  & \textbf{(12.00, 1.00)} & \textbf{(1.20, 1.00)} \\
\textbf{Qwen3-30B-T}  & \textbf{0.51} & \textbf{-0.35} & \textbf{(37.95, 0.10)} & \textbf{(1.20, 1.00)} \\
Qwen3-4B-T            & 0.51 & -0.39 & (12.00, 1.00) & (1.20, 1.00) \\
Phi-4-mini-R          & 0.43 & -0.35 & (12.00, 1.00) & (1.20, 1.00) \\
R1-Qwen-14B           & 0.41 & -0.32 & (12.00, 1.00) & (1.20, 1.00) \\
\textbf{R1-Llama-70B} & \textbf{0.36} & \textbf{-0.25} & \textbf{(33.84, 0.87)} & \textbf{(1.20, 1.00)} \\
R1-Llama-8B           & 0.29 & -0.27 & (12.00, 1.00) & (1.20, 1.00) \\
\bottomrule
\end{tabular}
\end{table}

\section{Segmentation Prompt and NLI Scorer Input}
In this section we provide the prompt we used to segment reasoning chains into segments $\mathcal{V} = \{v_1, \dots, v_n\}$ and the design of NLI-Scorer inputs.

\label{appendix_c_prompt}

\begin{tcolorbox}[
  colback=gray!5, colframe=black!20, boxrule=0.5pt, arc=2pt,
  left=6pt, right=6pt, top=6pt, bottom=6pt
]
\small
\textbf{Prompt used for indexed segmentation}

\medskip
You segment reasoning traces into logical blocks for argument-graph analysis. The
reasoning trace has already been split into numbered units using a rule-based
preprocessor. Return only compact segment ranges over those numbered units. Return
exactly one valid JSON object with this schema and nothing else:

\begin{lstlisting}[basicstyle=\ttfamily\scriptsize]
{ "segment_ranges": [[start_unit_index, end_unit_index], ...] }
\end{lstlisting}

\textbf{Rules:}
\begin{enumerate}[leftmargin=*, itemsep=1pt, topsep=2pt, label=\arabic*.]
  \item Do not rewrite, paraphrase, or quote the reasoning text.
  \item Each range is 1-based and inclusive, referring to the numbered units below.
  \item Ranges must be contiguous, non-overlapping, and cover every unit exactly once.
  \item Group consecutive units that serve the same immediate reasoning purpose into one segment.
  \item Start a new segment when the reasoning changes function (setup, sub-conclusion,
        computation, option evaluation, final answer, etc.).
  \item Keep closely related false starts or self-corrections together if they resolve the same local issue.
  \item If uncertainty resolves into a conclusion used in the next step, start a new segment at that boundary.
  \item Keep brief scene-setting with the reasoning move it introduces unless it is clearly its own step.
  \item A segment may span multiple units forming one coherent logical move.
  \item Use as many segments as the reasoning naturally requires.
  \item No markdown, no code fences, no extra keys, no prose outside the JSON.
\end{enumerate}

\medskip
\textbf{Reasoning trace to segment:} \texttt{[1] \ldots\ [2] \ldots}
\end{tcolorbox}
As mentioned earlier, segmentation by an instruct model can be substituted by a simple deterministic split without losing significant calibration quality. We can simply divide the trace at blank lines (the regular expression
\texttt{\textbackslash n\textbackslash s*\textbackslash n+}, i.e.\ two or more
consecutive line breaks), yielding one segment per paragraph.

We designed the input to our NLI-Scorer as follows:
\begin{tcolorbox}[
  colback=gray!5, colframe=black!20, boxrule=0.5pt, arc=2pt,
  left=6pt, right=6pt, top=6pt, bottom=6pt
]
\small
\textbf{Input format for NLI edge scoring}

\medskip

\medskip
\textbf{Segment--segment edge $(v_i, v_j)$:}
\begin{description}[leftmargin=7em, style=nextline, itemsep=1pt, topsep=2pt]
  \item[Premise] text of segment $v_i$
  \item[Hypothesis] text of segment $v_j$
\end{description}

\textbf{Segment--option edge $(v_i, h_k)$:}
\begin{description}[leftmargin=7em, style=nextline, itemsep=1pt, topsep=2pt]
  \item[Premise] text of segment $v_i$
  \item[Hypothesis] \texttt{The answer is Option X: [option text]}
\end{description}

\medskip
\textbf{Example} (segment $v_3$ of Fig.~\ref{fig:edge_stats}c against option A):
\begin{description}[leftmargin=7em, style=nextline, itemsep=1pt, topsep=2pt]
  \item[Premise] \texttt{The Skyrmion is a topological excitation, not a Goldstone boson. So the answer is Skyrmion, choice A.}
  \item[Hypothesis] \texttt{The answer is Option A: Skyrmion}
  \item[Output] $(p_E, p_N, p_C) = (0.98,\, 0.02,\, 0.00)$
</parameter>
\end{description}
\end{tcolorbox}

\section{Computing Infrastructure}\label{sec:computing_infra}
Experiments were run on 2 NVIDIA H200 GPUs.



\end{document}